\documentclass[preprint,3p,twocolumn]{elsarticle}
\usepackage{lineno,hyperref,hhline}
\usepackage{caption}
\usepackage{subcaption}
\usepackage{amsfonts,graphicx,amsmath}
\usepackage{multicol,multirow,longtable,setspace}
\usepackage{adjustbox}

\journal{Artificial Intelligence in Medicine}

\usepackage[obeyFinal]{todonotes}

\begin{document}

\begin{frontmatter}

\title{Early detection of sepsis utilizing deep learning on
electronic health record event sequences}

\author[Enversion,two]{S. M. Lauritsen\corref{correspondingauthor}}
\cortext[correspondingauthor]{Corresponding author Email: sla@enversion.dk}
\author[Enversion]{M. E. Kal{\o}r}
\author[Enversion]{E. L. Kongsgaard}
\author[two,endo]{K. M. Lauritsen}
\author[four]{M. J. J{\o}rgensen}
\author[two,four]{J. Lange}
\author[Enversion,three]{B. Thiesson}

\address[Enversion]{Enversion A/S}
\address[two]{Department of Clinical Medicine, Aarhus University, Denmark}
\address[three]{Department of Engineering, Aarhus University School of Engineering, Denmark}
\address[four]{Horsens Regional Hospital, Denmark}
\address[endo]{Department of Endocrinology and Internal Medicine, Aarhus University Hospital, Aarhus, Denmark}

\begin{abstract}
\textbf{Background:} The timeliness of detection of a sepsis incidence in progress is a crucial factor in the outcome for the patient. Machine learning models built from data in electronic health records can be used as an effective tool for improving this timeliness, but so far the potential for clinical implementations has been largely limited to studies in intensive care units. This study will employ a richer data set that will expand the applicability of these models beyond intensive care units. Furthermore, we will circumvent several important limitations that have been found in the literature: 1) Models are evaluated shortly before sepsis onset without considering interventions already initiated. 2) Machine learning models are built on a restricted set of clinical parameters, which are not necessarily measured in all departments. 3) Model performance is limited by current knowledge of sepsis, as feature interactions and time dependencies are hard-coded into the model.
\textbf{Methods:}
In this study, we present a model to overcome these shortcomings using a deep learning approach on a diverse multicenter data set. We used retrospective data from multiple Danish hospitals over a seven-year period. Our sepsis detection system is constructed as a combination of a convolutional neural network and a long short-term memory network. We suggest a retrospective assessment of interventions by looking at intravenous antibiotics and blood cultures preceding the prediction time.
\textbf{Results:}
Results show performance ranging from AUROC 0.856 (3 hours before sepsis onset) to AUROC 0.756 (24 hours before sepsis onset). Evaluating the clinical utility of the model, we find that a large proportion of septic patients did not receive antibiotic treatment or blood culture at the time of the sepsis prediction, and the model could therefore facilitate such interventions at an earlier point in time.
\textbf{Conclusion:}
We present a deep learning system for early detection of sepsis that is able to learn characteristics of the key factors and interactions from the raw event sequence data itself, without relying on a labor-intensive feature extraction work. Our system outperforms baseline models, such as gradient boosting, which rely on specific data elements and therefore suffer from many missing values in our dataset.

\end{abstract}

\begin{keyword}
Sepsis, Clinical decision support systems, Machine learning, Medical informatics, Early diagnosis, Electronic health records
\end{keyword}

\end{frontmatter}

\section{Introduction}
Sepsis is one of the most common causes of death globally \cite{peterA2018}. The World Health Organization estimates that more than six million people die of sepsis annually, and many of these deaths are preventable \cite{SecretariaW2017}. In the United States, severe sepsis affects more than 700,000 patients each year at a cost of more than 20 billion dollars \cite{Murphy_SL_et_al_2013,Derek_C._Angus_et_al_2001}. Early detection of sepsis has shown to improve patient outcomes, but it remains a challenging problem in medicine \cite{Jean-Louis_Vincent2016}. Even experienced physicians have difficulties in detecting sepsis early and accurately, as the early symptoms associated with sepsis may also be caused by many other clinical conditions \cite{Jones_AE_et_al_2010}. Previous studies have shown that machine learning (ML) models trained from data in individual patient electronic health records (EHR) may be used for the early detection of sepsis 
\cite{Joseph_Futoma_et_al_2017AI,Steven_Horng_et_al_2017,Joseph_Futoma_et_al_2017,Qingqing_Mao_et_al_2018,Shamim_Nemati_et_al_2018,David_W_Shimabukuro_et_al_2017}. The ML models for sepsis detection far exceed the predictive ability of existing clinical early warning system scores, such as the National Early Warning Score (NEWS)\cite{Joseph_Futoma_et_al_2017AI,Joseph_Futoma_et_al_2017,Qingqing_Mao_et_al_2018,David_W_Shimabukuro_et_al_2017,Calvert_JS_et_al_2016,Md._Mohaimenul_Islam_et_al_2019}. Recently, Shimabukuro et al. demonstrated several positive effects with the use of an ML model for sepsis detection in a randomized trial. Usage of the model led to an in-hospital mortality decrease of 12.4 percentage points (p=0.018) and an average length of stay decrease from 13.0 to 10.3 days (p=0.042) \cite{David_W_Shimabukuro_et_al_2017}. However, the current studies have limitations. First, most studies build their ML models on a limited set of clinical parameters, such as vital signs, which must be collected at the hospital department before the model can be used. Emergency departments and intensive care units (ICU) often have guidelines for frequent registration of vital signs, but this is typically not the case in many medical and surgical departments. The applicability and deployment potential of the models is therefore limited due to the comprehensive data registration requirements that are imposed on the departments in which the models are to be used. Second, during model evaluation, it is customary to report only receiver operating characteristic (ROC) curves and the derived area under the ROC curve (AUROC). This type of evaluation is chosen in spite of claims that AUROC is purely a measure of predictive ability and does not measure expected clinical usefulness, as it does not take prevalence into account \cite{Steve_Halligan_et_al_2015,Kevin_McGeechan_et_al_2014,Talluri_R_Shete_S2016}. AUROC may be misleading when applied to data sets with a high imbalance between positive and negative samples, which is often the case within the field of health science. Additionally, most studies are evaluated by ROC curves at a fixed time relative to the time of sepsis onset. In a real clinical setting, the evaluation should start at the time the patient arrives at the hospital, and the algorithm should be used for inference multiple times thereafter. Finally, the clinical utility of the models is typically not investigated in relation to potential interventions. As an example, it is not reported whether sepsis treatment has already been initiated at the time of the early detection. 
\par

In this paper we present 1) a scalable deep learning \cite{Yann_LeCun_et_al_2015} approach for early sepsis detection on the heterogeneous data set present outside ICUs. 2) We suggest a sequence evaluation approach that provides realistic estimations of model performance. 3) We evaluate the clinical utility of the model in relation to early interventions with blood cultures and antibiotics.
\par

\section{Materials and methods}
\subsection{Data population and data sources}
The data included health data on all citizens 18 years or older with residency in one of four Danish municipalities (Odder, Hedensted, Skanderborg, and Horsens). We used data from the secondary health sector in combination with nationwide registers for the period 2010 to 2017. The data from the secondary health sector contained information from the EHR, including biochemistry, medicine, microbiology, medical imaging, and the patient administration system (PAS). The data constituted raw health events of a sequential nature and were used in the main features for the ML models, as described in more detail in Sections \ref{data_representation} - \ref{data_preporcession}. 
\par

\begin{table}[!h]
\adjustbox{max width=\linewidth}{%
 			\centering
\begin{tabular}{p{0.3\linewidth}p{0.5\linewidth}}
\hline
\multicolumn{1}{p{0.3\linewidth}}{ \textbf{Source system}} & 
\multicolumn{1}{p{0.9\linewidth}}{ \textbf{Data type}} \\
\hhline{--}
\multicolumn{1}{p{0.3\linewidth}}{ Electronic Health Record (Patient \newline administration system)} & 
\multicolumn{1}{p{0.9\linewidth}}{ Diagnoses (International classification disease - 10; ICD-10), procedures (NCSP -- the NOMESCO Classification of Surgical Procedures), booking information, health content (structured notes containing physiological measurements, symptom classifications, check box data such as smoking and exercise habits) } \\
\hhline{--}
\multicolumn{1}{p{0.3\linewidth}}{ Electronic Health Record (Medication module)} & 
\multicolumn{1}{p{0.9\linewidth}}{ Dates and times for prescriptions and dispensing together with information on ingredients, dose, administration routes.} \\
\hhline{--}
\multicolumn{1}{p{0.3\linewidth}}{ Laboratory system} & 
\multicolumn{1}{p{0.9\linewidth}}{ Microbiology and blood gas analysis} \\
\hhline{--}
\multicolumn{1}{p{0.3\linewidth}}{ Medical imaging system} & 
\multicolumn{1}{p{0.9\linewidth}}{ Image descriptions from computed tomography, magnetic resonance imaging, ultrasound, X-ray, positron-emission tomography} \\
\hhline{--}
\multicolumn{1}{p{0.3\linewidth}}{ National Patient \newline Register} & 
\multicolumn{1}{p{0.9\linewidth}}{ Hospital admissions, diagnoses (ICD-10), procedures (NCSP)} \\
\hhline{--}
\multicolumn{1}{p{0.3\linewidth}}{ Civil Registration \newline System} & 
\multicolumn{1}{p{0.9\linewidth}}{ Patient demographics: age, address, and marital status.} \\
\hhline{--}

\end{tabular}
}
\caption{Data sources}
\label{table1}
 \end{table}

\begin{figure*}[!h]
	\begin{center}
		\includegraphics[width=\textwidth]{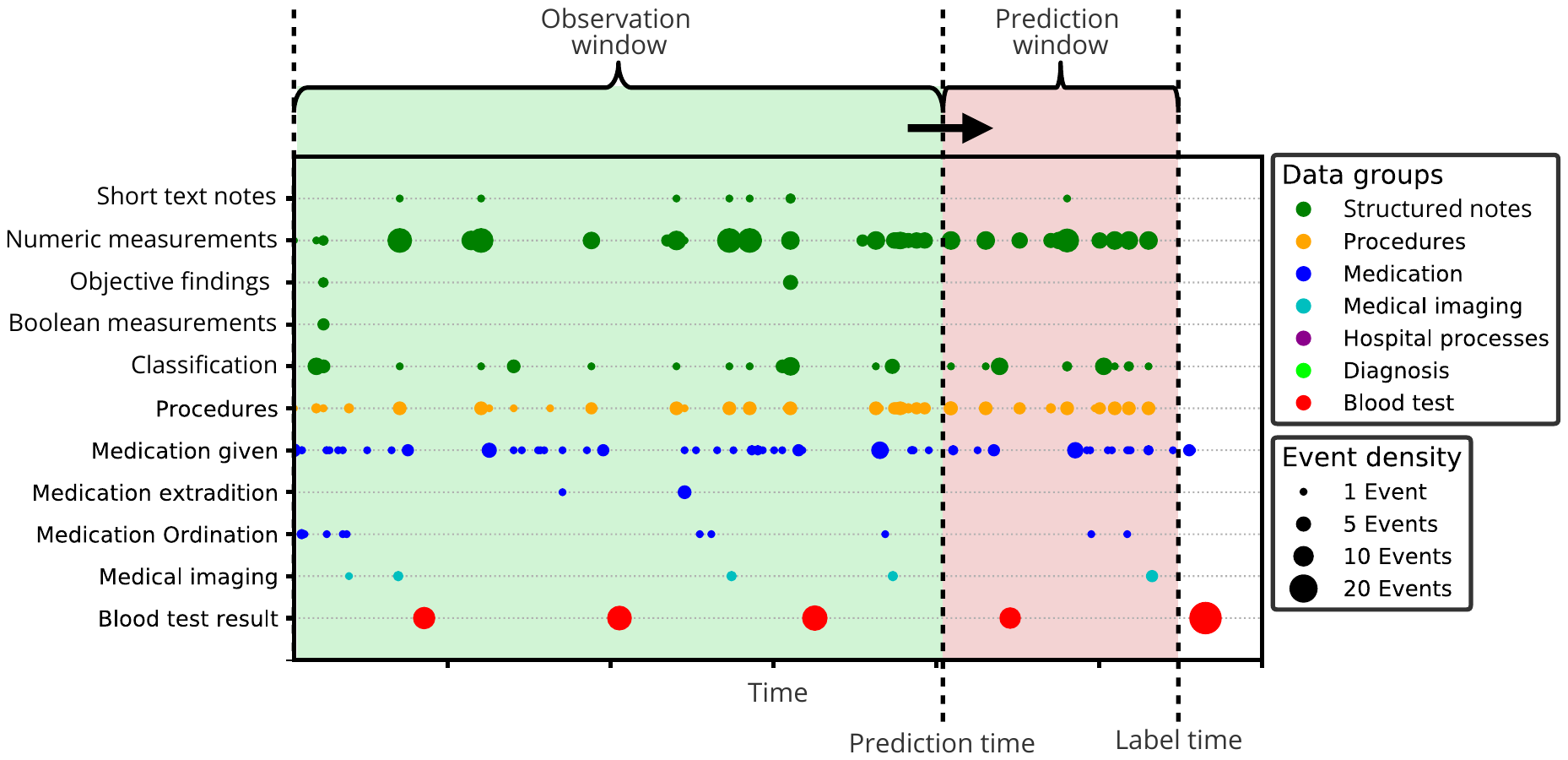}
		\setlength{\belowcaptionskip}{-12pt}		
		\caption{A visual example of data for a randomly chosen sepsis patient. The observation window has a green background while the prediction window has a red background. The transition between the two windows is called the prediction time. The prediction time is not static, as displayed in this snapshot; instead, it is shifting from the beginning of each sequence to the label time, as the hospitalization progresses. For the sepsis-positive sequences, the label time corresponded to the time of sepsis onset. For the negative sequences, the label time was randomly chosen within the admission.}
		\label{fig:__A_visual_example}
	\end{center}
\end{figure*}

The National Patient Register \cite{Morten_Schmidt_et_al_2015} and the Civil Registration System \cite{Carsten_Bcker_Pedersen2011} contain information of a more contextual type, such as previously registered diagnoses, procedures, hospital admissions, marital status, and housing situation. These data were used to additionally include contextual covariates with information about comorbidities, age, and marital status, which had been registered preceding the current admission. Table \ref{table1} shows a list of the data sources that were used in the study. The data were extracted from the research project ``CROSS-TRACKS''\footnote{http://www.tvaerspor.dk/}. Representation of the raw sequential data for a random patient is shown in Figure \ref{fig:__A_visual_example}. The different data sources are color coded so that events from the same data source have the same color. For example, both given and prescribed medications are colored blue. The circle size indicates how many events have been observed with the same timestamp. The size of a circle scales with the square root of the number of concurring events. Four circle examples of 1, 5, 10, and 20 events are shown.
\par


\subsection{Inclusion criteria and dataset preparation  }
\label{inclusion_criteria_and_dataset}
\begin{figure}[!h]
	\begin{center}
		\includegraphics[width=\linewidth,height=3in]{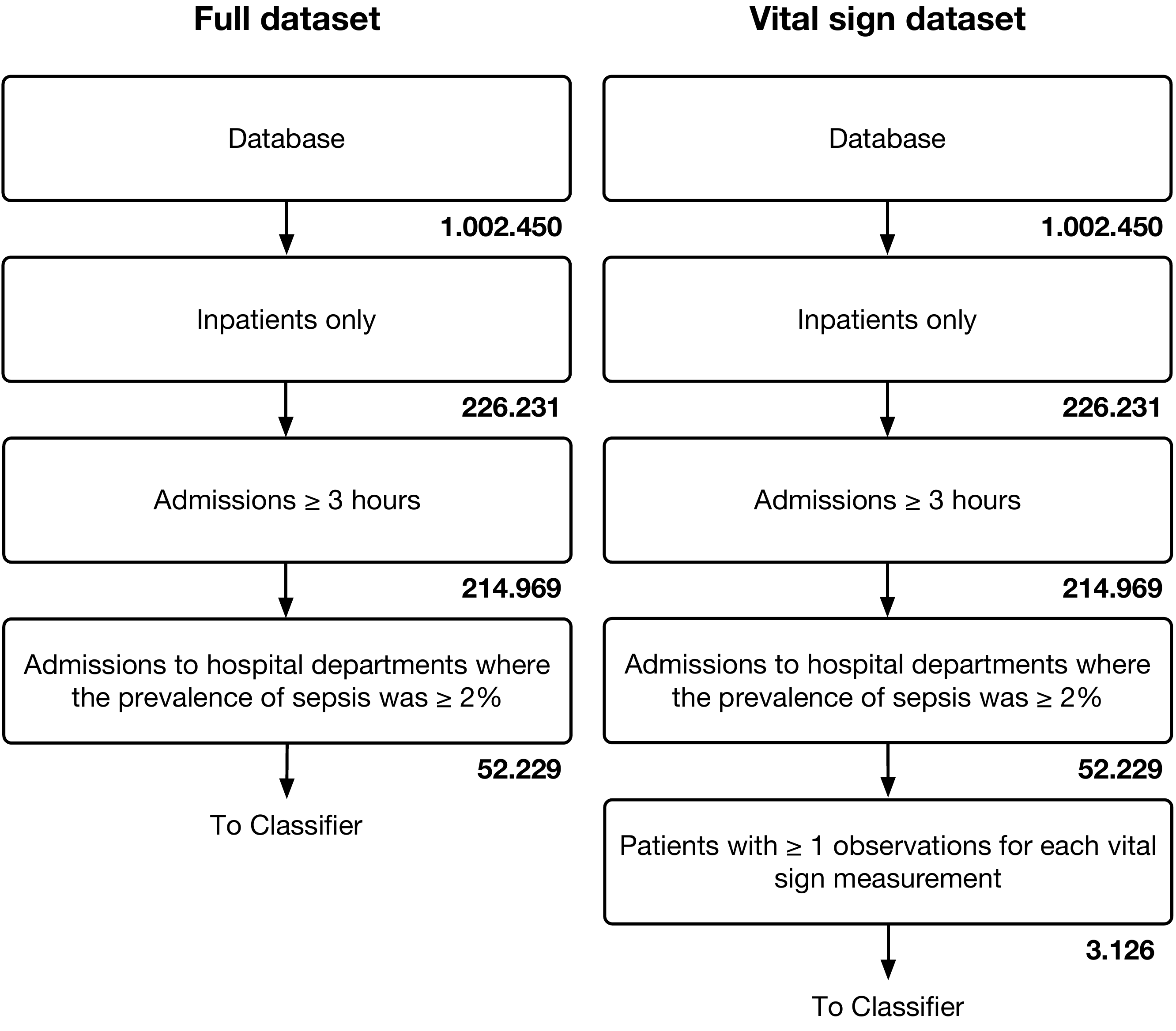}
		\setlength{\belowcaptionskip}{-12pt}		
		\caption{Inclusion flow chart Left) inclusion criteria for the Full dataset. Right) inclusion criteria for the Vital sign data set.}
		\label{fig:__Inclusion_flow_chart}
	\end{center}

\end{figure}

The data set was constructed in the following stepwise manner. First, all relevant hospital contacts were identified from a set of 1,002,450 contacts. From this set, 776,219 outpatient contacts were removed, leaving 226,231 inpatient contacts to be considered. Second, 11,262 admissions with a duration of less than three hours were removed. Third, admissions to hospital departments where the overall prevalence of sepsis was less than 2$\%$  (162.740 contacts) were removed in order to limit the number of false positives, leaving 52,229 contacts in total in the data set). In addition, a second data set, called the vital sign data set, was constructed by removing 49,000 contacts with incomplete vital sign measurements (systolic blood pressure, diastolic blood pressure, heart rate, respiratory rate, peripheral capillary oxygen saturation, and temperature) in the three hours preceding the label time. The vital sign data set thus included a total of 3,126 contacts (Figure \ref{fig:__Inclusion_flow_chart}). This was done to have a data set with the possibility of direct comparison between all models.


For both data sets, hospital contacts that led to sepsis were identified (see section \ref{target_definition_for_the} below), and the data were split into two parts: one with sepsis-positive contacts and one with sepsis-negative contacts. Sepsis-positive contacts were further divided into training data (80$\%$), validation data (10$\%$), and test data (10$\%$). The training data were used to fit the model parameters. The validation data were used to perform an unbiased evaluation of a model fit during training, and the test data were used to provide an unbiased evaluation of the final model fit on the training data. In the training data, the sepsis positive contacts were oversampled by a factor of ten. 
\par

For each sequence, we considered at most five days of data prior to the label time, such that the sum of the observation and prediction window was five days (Figure \ref{fig:__A_visual_example}). For the positive sequences, the label time corresponded to the time the positive label was obtained (sepsis onset). For the negative sequences, there was no obvious label time, so we selected a pseudo-random time during the admission, excluding the first and last three hours of the admission. In Figure \ref{fig:__A_visual_example}, the observation window is shown with green and the prediction window is shown with red. The transition between the two windows is marked by the prediction time, which is not static but rather sliding from the beginning of each sequence to the label time. In this way, both windows are changing size as the prediction time shifts forward in real time. The label time will be referred to as \textit{t}. Thus, the time three hours before the label time can be written as t-3 hours. Only 5,9$\%$  of patients in the test data set have a registration of at least two of the vital sign measurements three hours before the label time. Since several of the selected evaluation metrics are sensitive to stratification and prevalence, we have chosen to test our models on both the full data set and the vital sign data set. This allows the models to be compared directly for this smaller sample of patients.
\par

\subsection{Target definition for the early detection of sepsis}\label{target_definition_for_the}

After the inclusion of hospital admissions, each admission underwent a binary classification process to denote it as either \textit{sepsis-positive} or \textit{sepsis-negative}. The classification was made based on patients meeting the gold standard for sepsis, which is based on the 2001 consensus definition of sepsis. \cite{Mitchell_M._Levy2003}: 
\par

The presence of two or more Systemic Inflammatory Response Syndrome (SIRS) criteria paired with a suspicion of infection$"$. SIRS criteria are defined as:
\begin{itemize}
	\item heart rate $\textgreater$ 90 beats/min\par

	\item body temperature $\textgreater$ 38$ ^{\circ} $ C or $\textless$ 36$ ^{\circ} $ C\par

	\item respiratory rate $\textgreater$ 20 breaths/min or PaCO2 (alveolar carbon dioxide tension) $\textless$ 32 mm Hg\par

	\item white cell count $\textgreater$ 12 x 109 cells/L or $\textless$ 4 x 109 cells/L
\end{itemize}\par

A hospital admission was labeled sepsis-positive if the EHR contained data registration fulfilling the 2001 consensus sepsis definition during admission. Otherwise, the hospital admission was labeled sepsis-negative. Importantly, the fulfillment of the 2001 consensus sepsis definition is an independent EHR registration that occurs continually during admission. This registration is in contrast to the registration of diagnoses, which often relates to the time of discharge. Hence, in this study, the gold standard may be fulfilled and registered even though vital sign measurements have not yet been entered into the EHR.

\par

\subsection{Data representation}
\label{data_representation}
In the raw data, each sample represents a given patient as a time-ordered sequence of EHR events  \( E= \left( e_{1},e_{2}, \ldots,e_{T} \right)  \), where  \( e_{t} \)  is an observed event ordered by  \( t \in 1, \ldots,T \). Each event consists of three elements: a time stamp, an event category (e.g., blood pressure or medication code), and a value. The time for event  \( e_{t} \)  is denoted as  \( t \left( e_{t} \right)  \), the category \(  c \left( e_{t} \right)  \), and the value  \( v \left( e_{t} \right)  \). For example, if the category  \( c \left( e_{t} \right)  \)  is blood pressure, then  \( v \left( e_{t} \right)  \in \mathbb{R}^{2} \), as it contains both the systolic and diastolic measurements. Notice that only for the sequential neural network model, the detailed ordering of events is important. In Figure \ref{fig:__A_visual_example}, a visual example of a time-ordered sequence of EHR events is given. 
\par

The raw event data are transformed through two-step vectorization of the individual events.\  The first step will represent each event  \( e_{t} \)  by a very sparse vector  \( e_{t}  \) with an entry for all event-value types that can be observed across all patients. The size of this vector will be greater than the number of different event categories, as a category may have more than one measurement (e.g., for the blood pressure event from above, the event vector  \( e_{t}  \)  will have two nonzero elements, one for each measurement in  \( v \left( e_{t} \right)  \)). In the second step, each vector entry is further transformed as follows. Categorical features are converted into their corresponding one-hot binary feature vector, numerical features are standard normalized, and hierarchical features, such as diagnosis codes, are represented as multi-hot vectors with an entry in each of the present levels of the diagnosis hierarchy. The resulting vectorization of a given event  \( e_{t} \)  is therefore a very sparse, but not necessarily one-hot, vector  \( e_{t} \)  of size 80,000. 
\par

A raw event vector sequence may be partitioned into intervals of time, where the raw event vectors are then aggregated within a time interval  \( I \). We will let  \( e_{I} \) \textbf{ } denote the interval aggregation of all  \( e_{t} \)  where  \( t \left( e_{t} \right)  \in  I \). Different aggregation functions are applied across different elements in the event vectors. Binary (categorical) outcomes, such as procedure codes, are aggregated to numerical counts, and numerical measurements, such as blood pressure, are converted to minimum, maximum, and mean values. Naturally, depending on the degree of aggregation, the ordering of the events in the entire event sequence is ignored to a greater or lesser extent.
\par

Similar to the raw event vector, we also construct a vectorization of the context for a given patient. That is, meta data, such as demographics and the patient's comorbidities prior to the first raw event, are considered in a model. The context vector is denoted by  \( c \)  and is, in contrast to the raw event vectors  \( e_{t} \), not dependent on the sequence ordering in the model.
\par

\par

\subsection{Data preprocessing and model design }
\label{data_preporcession}

The models in this study were built with an onset using three different approaches: 1) a classical epidemiological approach, where the model includes a small group of selected and clinically well-founded features; 2) a more data-driven approach, where all the available data are used in a slightly aggregated form to train a non-sequential neural network; and finally 3) a data-driven approach, where the available data are used in their sequenced form for the training of a sequential neural network. The first two models serve as baseline comparison models.
\par

\subsubsection{Gradient boosting }

In the simplest baseline model, called GB-Vital, we replicate a well-known sepsis detection model from the literature, which has shown excellent results in a randomized study \cite{David_W_Shimabukuro_et_al_2017}. The full technical description of the model can be found in \cite{Qingqing_Mao_et_al_2018}. The explanatory features for this model are constructed by considering only six vital-sign events from the raw EHR event sequences: systolic blood pressure, diastolic blood pressure, heart rate, respiratory rate, peripheral capillary oxygen saturation, and temperature. The constructed features highly aggregate the sequence information, and only limited ordering information is retained. That is, for each of the six vital signs, five features are constructed to represent the average value for the current hour, the prior hour, and the hour prior to that hour along with the trend value between two succeeding hours. 
\par
Based on these 30 features (five values from each of the six measurement channels), the GB-Vital model is constructed as a gradient boosted classifier of decision trees. As in \cite{Qingqing_Mao_et_al_2018}, each tree in the gradient boosting model is limited to split at most six times, and no more than 1000 trees are aggregated to generate the risk prediction. The model was trained in Python using the Gradient Boosting Classifier in the Scikit-learn package.
\par

\subsubsection{Multilayer perceptron}
In the more advanced baseline model, we constructed a standard multilayer feedforward neural network in the form of a multilayer perceptron (MLP). The model does not limit data to include only vital signs. Instead, features were constructed by aggregating entire event vectors in  \( E \)  across retrospective windows of time, including intervals of 1 hour, 2 hours, 4 hours, 8 hours, 16 hours, and 32 hours preceding the time of the label. Notice that with this coarse aggregation of events, the ordering of the events is basically ignored, except for the fact that the final feature vector for the model concatenates the aggregating event vectors in  \( E \)  across the different sized windows of time. Finally, to remove noise and reduce dimensionality, we only consider features that are present in at least 100 sample sequences of the training data, resulting in a reduction from approximately 100,000 to 5000 distinct entries in the feature vector for each retrospective timespan, or approximately 30,000 features in total. 
\par

The model structure for the MLP in this study feeds the 30,000 input units together with the 26 contextual features into two hidden layers of 200 units each, which is then followed by the binary decision. The model was trained to optimize the cross entropy loss using the Adam optimizer \cite{Diederik_P._Kingma_Jimmy_Ba2014} with mini-batches of size 50, a learning rate of 0.0001, and a dropout of 30$\%$  to prevent overfitting. Keras 2.2.2 with a TensorFlow 1.11 backend was used for the MLP experiments in this study.
\par

\subsubsection{CNN-LSTM}
In this model we considered all elements in the entire event sequence  \( E \) for a patient. As with the MLP model, we only considered events that were present in at least 100 sample sequences of the training data, resulting in approximately 5000 distinct entries in the event vector representation  \( e_{t} \)  for each event  \( e_{t} \).
\par

The event sequences are further pre-processed by 1) a temporal preserving aggregation step, 2) a gap-filling step, and 3) a context concatenation step. In the temporal preserving aggregation step, all event vectors are grouped into five-minute non-overlapping blocks  \( B \subseteq E \) such that the maximum time between two events in each block is five minutes. The main reason for this step is to reduce the number of inputs to the model in order to improve computational efficiency. However, the temporal aggregation also has the effect of discarding the order of events within each five-minute block, which is arguably determined by the order in which the healthcare professional enters the information into the EHR rather than by the true order of events that close in time. The gap-filling step now fills the sequence with empty event vectors such that all feature vectors in the sequence are equidistant in time. In this way, there is a vector for every five minutes in the sequence, some of which are aggregations of one or more event vectors while the remainder are empty vectors. The sequence can therefore be represented as a sparse matrix of shape  \( N \times K \), where  \( N \)  is the number of five-minute vectors in the longest sequence (\( N= \left[ \frac{t \left( e_{T} \right) -t \left( e_{1} \right) }{5~minutes} \right]  \)) and  \( K \)  is the number of entries in the event feature vectors  \(  ( K \approx 5000 ) \). Finally, each of the  \( N \)  aggregated event vectors was concatenated with the same fixed context vector such that the final sequence matrix is of shape  \( N \times  \left( K+C \right)  \), where C is the number of entries in the context vector  \(  \left( C \approx 30 \right)  \). Recall that the context vector contains meta data about the patient, such as age, gender, and comorbidities. The intuition for concatenating the contextual data with every event is that the importance of certain EHR registrations may be supported by such contextual information.
\par

\begin{figure*}[!h]
	\begin{center}
		\includegraphics[width=\textwidth]{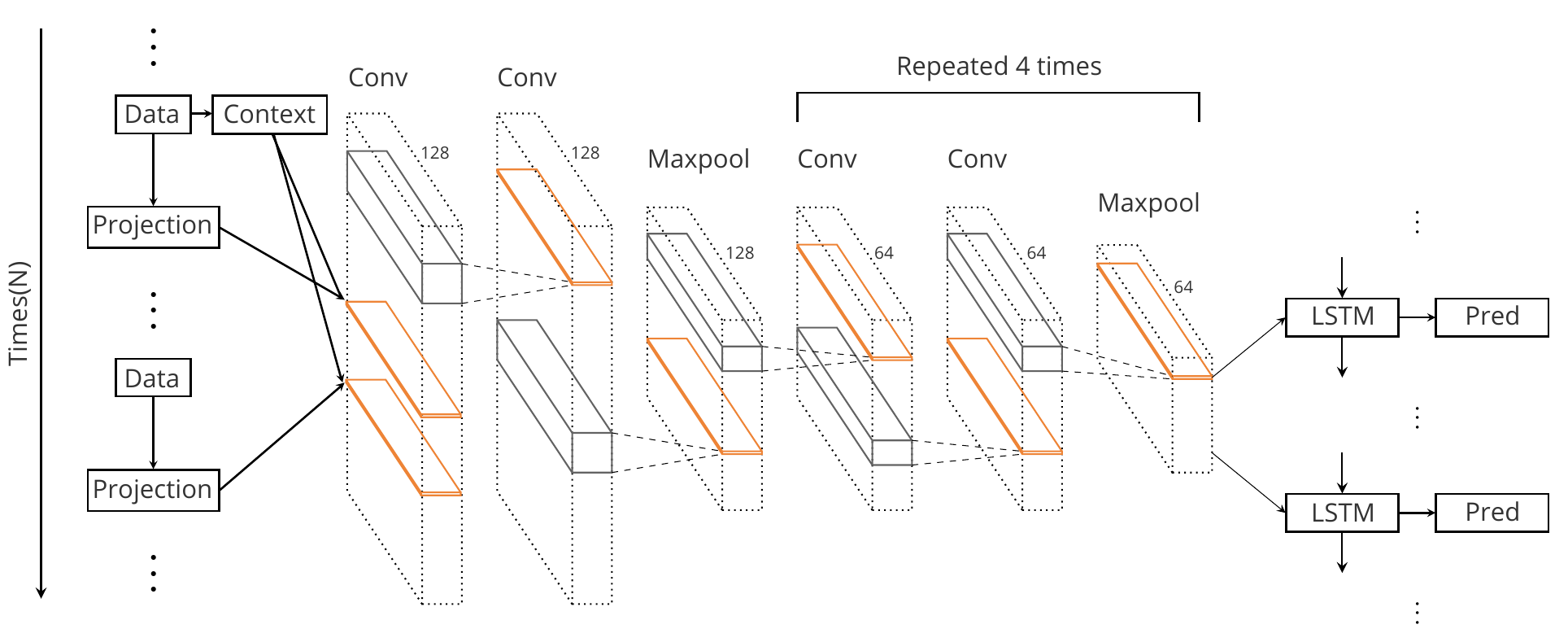}
		\setlength{\belowcaptionskip}{-12pt}
		\caption{The CNN-LSTM architecture for sepsis classification}
		\label{fig:__The_CNNLSTM_architecture_for_sepsis_classification}
	\end{center}
\end{figure*}

This classifier is structured as a convolutional neural network (CNN), followed by a recurrent layer of long short-term memory (LSTM) cells, also known as a CNN-LSTM model (or sometimes Long-term Recurrent Convolutional Network) \cite{Donahue_J_et_al_2017}. This architecture has been shown to learn robust temporal feature representations in the convolutional layers, which makes it easier for the LSTM layer to capture temporal dependencies compared to using the raw inputs \cite{Tara_N._Sainath_et_al_2015}.  The overall architecture of the classification model is illustrated in Figure \ref{fig:__The_CNNLSTM_architecture_for_sepsis_classification}. The model first projects the sparse inputs into dense 1000-dimensional vectors, reducing the dimensionality for the following convolutional layer by a factor of five. With inspiration from Conneau et al. \cite{Alexis_Conneau_et_al_2017}, short-term temporal developments for a patient are now captured in the model by a stack of $``$convolutional blocks$"$. A convolutional block consists of two one-dimensional ReLU-activated convolutional layers followed by a max-pooling layer. All convolutional layers have kernels of size 3, a stride of 1, and zero-padding is used. All max-pooling layers have a kernel size of 2 and a stride of 2, halving the temporal width of the input. To ensure that information across the convolutional blocks obeys the ordering of the input information without contaminating the output with information from the future, all kernels are causal in the sense that they only filter input from the current time and the past. 
\par


\par

\par

There are five convolutional blocks in the model. The initial block has a depth of 128 for both of the convolutional layers in the block, whereas the convolutional layers in the last four blocks all have a depth of 64. After the input filters through the five convolutional blocks, each temporal vector at the output contains partly overlapping information from a span of 15 hours and 30 minutes of the original input, and the temporal distance between each vector is 2 hours and 40 minutes. Finally, the model captures the long-term temporal development of a patient by allowing the output from the convolutional blocks to feed into an LSTM layer that incrementally builds up a representation of the temporal inputs and continually predicts an output. The LSTM layer has 64 units and is initialized with a random initial state. This layer consists of a $``$conventional$"$  LSTM layer with a forget gate, as defined in \cite{Alex_Graves2013}. Our experiments have showed that by adding the convolutional layers in front of the LSTM, we gain significant improvements in both efficiency and effectiveness compared to using a single stacked LSTM. The final prediction layer is a softmax layer.
\par
The model was trained on a NVIDIA Tesla V100 GPU. Convergence was reached after approximately 90 minutes. 
\par
\subsection{Model evaluation}
\label{model_evaluation}
The models in this study will produce a prediction value for each patient  \( p \)  that reflects the risk that a hospital admission may end up with sepsis if not intervened upon. This prediction will be in the range from zero to one, where the predicted risk should be higher for those patients at risk of later developing sepsis compared to those that are not. The discriminative power of a model with probabilistic predictions is typically evaluated at a range of decision thresholds  \( p_{\tau} \in  \left[ 0;1 \right]   \) for the binary decision  \( p>p_{\tau} \)  and reported in the form of receiver operating characteristic (ROC) curves, precision recall (PR) curves, area under ROC (AUROC), or mean average precision (mAP). We report these measures to enable easy comparison to existing and future studies that employ evaluations of this kind. 
\par
However, while discrimination is an important statistical property, it does not properly address clinical usefulness \cite{Steve_Halligan_et_al_2015,Kevin_McGeechan_et_al_2014,Talluri_R_Shete_S2016,Kevin_ten_Haaf_et_al_2017,Andrew_J._Vickers_Angel_M._Cronin2010,E._W._Steyerberg_Y._Vergouwe2014}. For example, if a false negative decision causes greater harm than a false positive decision, a model with high sensitivity may be preferable to a model with high specificity and lower sensitivity, although the latter model might have, say, a higher AUROC. In general terms, a model is clinically useful if the use of its decisions for patients leads to a better ratio between benefits and harms than not using the model. Grounded in the utility measure from the field of decision theory, decision curve analysis (DCA) assesses the clinical usefulness of a prediction model by evaluating the so-called net benefit (NB) at varying decision thresholds for the model (see, e.g., \cite{Andrew_J._Vickers_Elena_B._Elkin2006,Valentin_Rousson_Thomas_Zumbrunn2011}). Briefly, the net benefit is defined in Eq. (1), where a true positive and false positive are abbreviated as TP and FP, respectively:

    \begin{equation}
		\textrm{NB} = \frac{ \textrm{TP count} - (\textrm{FP count} \cdot \textrm{weight factor})} {\textrm{total count}}
	\end{equation}

Here, the weighting factor can be interpreted as the exchange ratio between the number of false positives that is acceptable in exchange for one true positive. This interpretation is important because it is informative of how the clinician weights the harm  \( H \)  of a false sepsis-positive decision over the benefit  \( B \)  of a true decision, with the rationale being to start intervention for a patient if the expected harm compared to the benefit is above the clinician's preference of exchange ratio  \( H/B \). For example, a weighting factor of  \( \frac{1}{10} \)  indicates that if the clinician misses one sepsis patient that could have been detected by the model, it is valued as being  \( 10 \)  times worse than unnecessarily classifying one healthy person to be at risk of sepsis.
\par

Although informative for the clinician, the formulation of the weighting factor as a harm--benefit ratio is not operational for the prediction models. Fortunately, a monotone one-to-one mapping to the decision threshold can be established, allowing a dual representation of the weighting factor as follows:

\begin{equation}
    \textrm{weighting factor} = \frac{H}{B} = \frac{p_{\tau}}{1-p_{\tau}}
\end{equation}

Given a model's computed risk prediction values for all patients, a decision curve in DCA can now be constructed by evaluating the net benefit for the binary decisions of opting in  \( p>p_{\tau} \)  or not  \( p \leq p_{\tau} \)  to the intervention across a range of different decision thresholds -- or equivalently, for a range of different harm--benefit exchange ratios.
\par

With different models to consider -– including the extreme options of never intervening or always intervening -- the clinician should favor the model with the highest net benefit at his personally determined H/B ratio. The curves allow the clinician to alter this ratio in the context of a given patient (e.g., in accordance with the patient's preferences). See Figure \ref{fig:6c} for an example of DCA curves. Finally, Eq. (2) can now be used to translate the identified  \( H/B \)  weighting factor into the following operating decision threshold in the chosen model
\begin{equation*}
 p_{\tau}=\frac{\textrm{weighting~factor}}{1+\textrm{weighting~factor}}
\end{equation*}

The attentive reader may have noticed that the above formulation of DCA exclusively focuses on the patients for whom an intervention will occur, as is the case in \cite{Andrew_J._Vickers_Elena_B._Elkin2006}. For the remaining patients, their hospital contact will continue as usual and not be affected by the prediction model. This is a more conservative evaluation than the formulation of DCA in \cite{Valentin_Rousson_Thomas_Zumbrunn2011}, where harms and benefits for non-interventions are also included. However, the latter DCA formulation is more demanding on the elicitation of the weighting factor to be used in an actual clinical setting, as it now relies on the clinician's ability to state a four-way relation between harms and benefits. In the following, we will use the former definition of net benefit in our DCA reporting.
\par


\begin{figure*}[!h]
	\begin{center}
		\includegraphics[width=\textwidth]{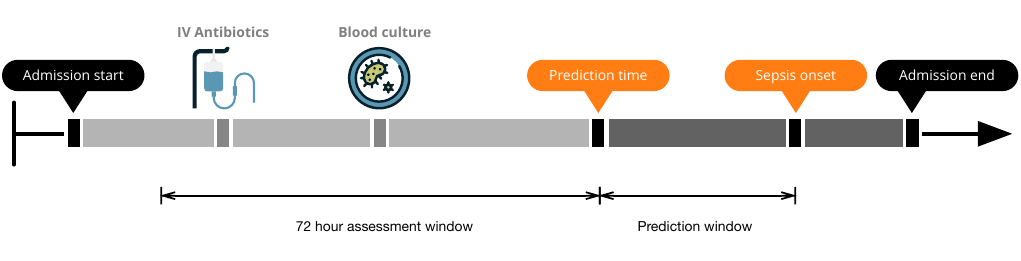}
		\setlength{\belowcaptionskip}{-12pt}
		\caption{Sequence evaluation with retrospective assessment of intervention potential (SERAIP). The figure shows the retrospective assessment of interventions looking at intravenous antibiotics and blood cultures preceding the prediction time.}
	\label{fig:sequance evaluation}
	\end{center}
\end{figure*}


\par

\par

\subsubsection{Sequence evaluation with retrospective assessment of intervention potential (SERAIP)}
The evaluation metrics described in Section \ref{model_evaluation} are generic and do not take into account how the ML model should be used in practice. If the evaluation is to express clinical utility in relation to usage, then two important aspects need to be addressed: 

\begin{itemize}
\item The sepsis onset time when the model will be used is unknown.
\item Potential interventions against sepsis may have already been initiated when the model will be used.
\end{itemize}

Concerning the first aspect: A model can perform very well when evaluated close to the onset of sepsis but at the same time perform critically bad when evaluated many hours before severe sepsis symptoms occur. When the model is used in a real-time clinical setting, the effect of a positive prediction will be an intervention that cannot be withdrawn. This implies that model performance in early timesteps must be carried through and accounted for when evaluating model performance in subsequent timesteps. We address the effect of previous model performance by defining the sequence prediction at time $t$,   \(p_{t}^{seq} \) as the maximum probability of all predictions until then. That is: 

\begin{equation}
{p_{t}^{seq}} = \mathop {\max }\limits_{0 \le t \le x} \left( p \right)
\end{equation}
where $p$ is the output probability, $t$ is the timestep variable, and $x$ is the timestep corresponding to the prediction time. In this way, a sepsis-positive classification will be maintained for the subsequent timesteps, as the effect of a positive prediction will be an intervention that cannot be withdrawn.

Concerning the second aspect: Predictions from an ML model are mostly useful if they can lead to actions. If interventions against sepsis are already initiated at the time of prediction, the model will not create any additional value. Generic evaluation metrics, such as ROC, PR, and DCA, do not measure the clinical usefulness of early sepsis detection in relation to potential interventions. 

We suggest adding a retrospective assessment of interventions to the evaluation by looking at intravenous antibiotics and blood cultures preceding the prediction time. We include registrations all the way back to 72 hours before the prediction time to ensure that all registrations related to clinical presumptions on sepsis are captured.

At each timestep this assessment is performed for all TP predictions, yielding the number of \textit{TP’s with intra venous  antibiotics} and \textit{TP’s with blood culture} individually and \textit{TP’s with intravenous antibiotics or blood culture} to get a measure of how many cases there are with a $p_{t}^{seq}$ above $\tau$ at prediction time that have a sepsis-related intervention. Finally, \textit{TP with no intervention} is reported  to indicate the potential for early intervention.

Intravenous antibiotics are identified as intravenous medications belonging to either the ATC J01 (antibacterial agents for systemic use) or ATC J02 (antimycobacterial agents) subgroups, and blood cultures are identified through the laboratory system for microbiology. A visual representation of this evaluation method named \textit{sequence evaluation with retrospective assessment of intervention potential} (SERAIP) is illustrated in Figure \ref{fig:sequance evaluation}
\par

\section{Results}
\begin{figure}[!h]
	\begin{center}
		\includegraphics[width=\linewidth,height=3in]{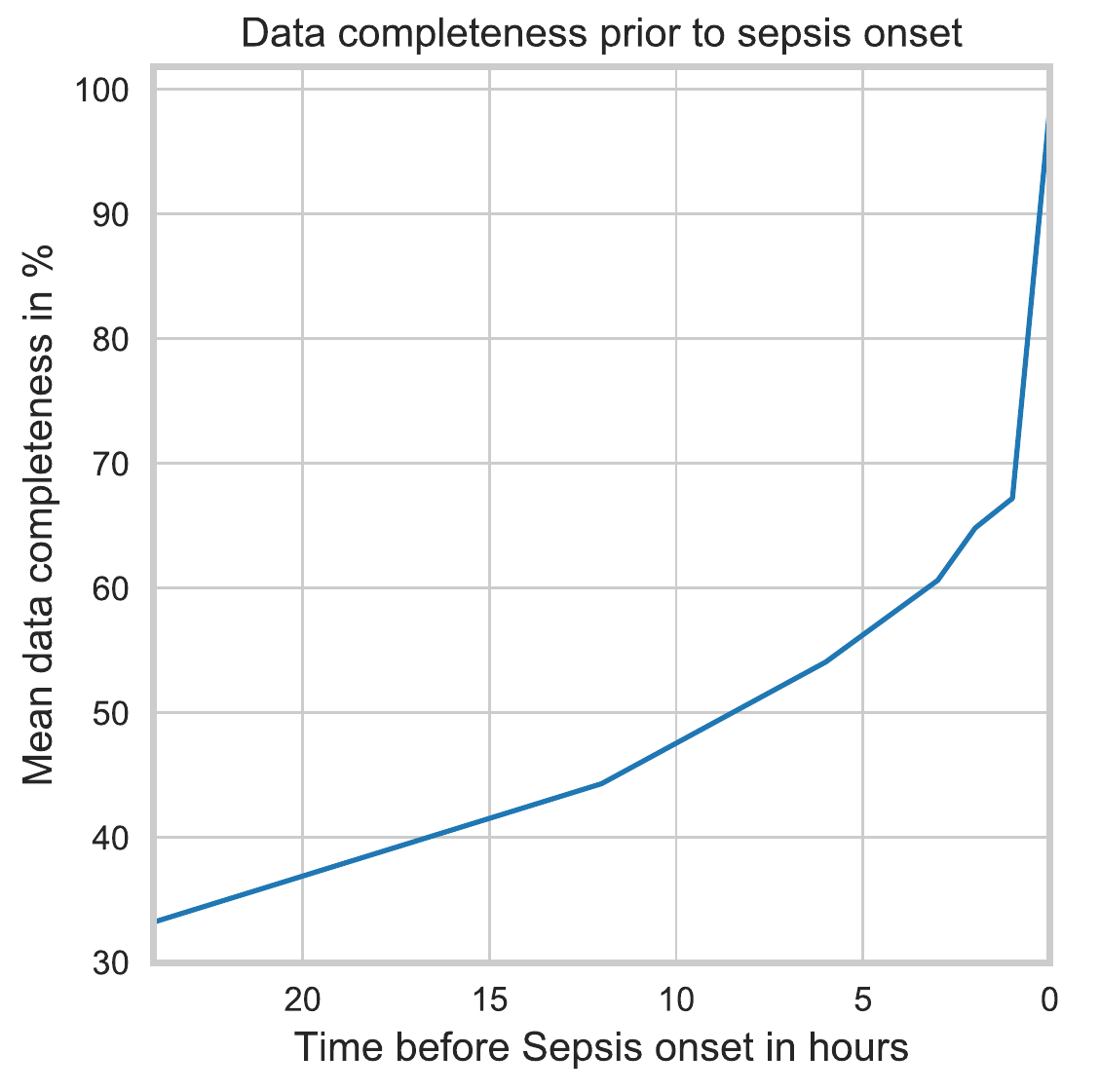}
		\setlength{\belowcaptionskip}{-12pt}		
		\caption{Data completeness of vital sign measurements decreased as a function of time before sepsis onset}
		\label{fig:__Meandata}
	\end{center}
\end{figure}

In our multi-center dataset, the data completeness of vital sign measurements decreased as a function of time before sepsis onset (Figure \ref{fig:__Meandata}).
In the interval from sepsis onset and six hours earlier in time ($t-6$ to $t-0$), two or more vital signs were registered for 100$\%$  of the septic patients. By moving the fixed size interval three hours back, covering the time from $t-9$ to $t-3$, the number decreased to 65$\%$. In the range from 18 hours before sepsis onset to 12 hours before sepsis onset ($t-18$ to $t-12$), the number was down to 43$\%$. Finally, in the interval between 24 to 30 hours before sepsis onset ($t-24$ to $t-30$), only 32$\%$  of the patients had registered two or more vital signs. 
\par


\subsection{Gradient Boosting}

\begin{figure*}[!h]
        \begin{subfigure}[b]{0.336\textwidth}
                \includegraphics[width=\textwidth]{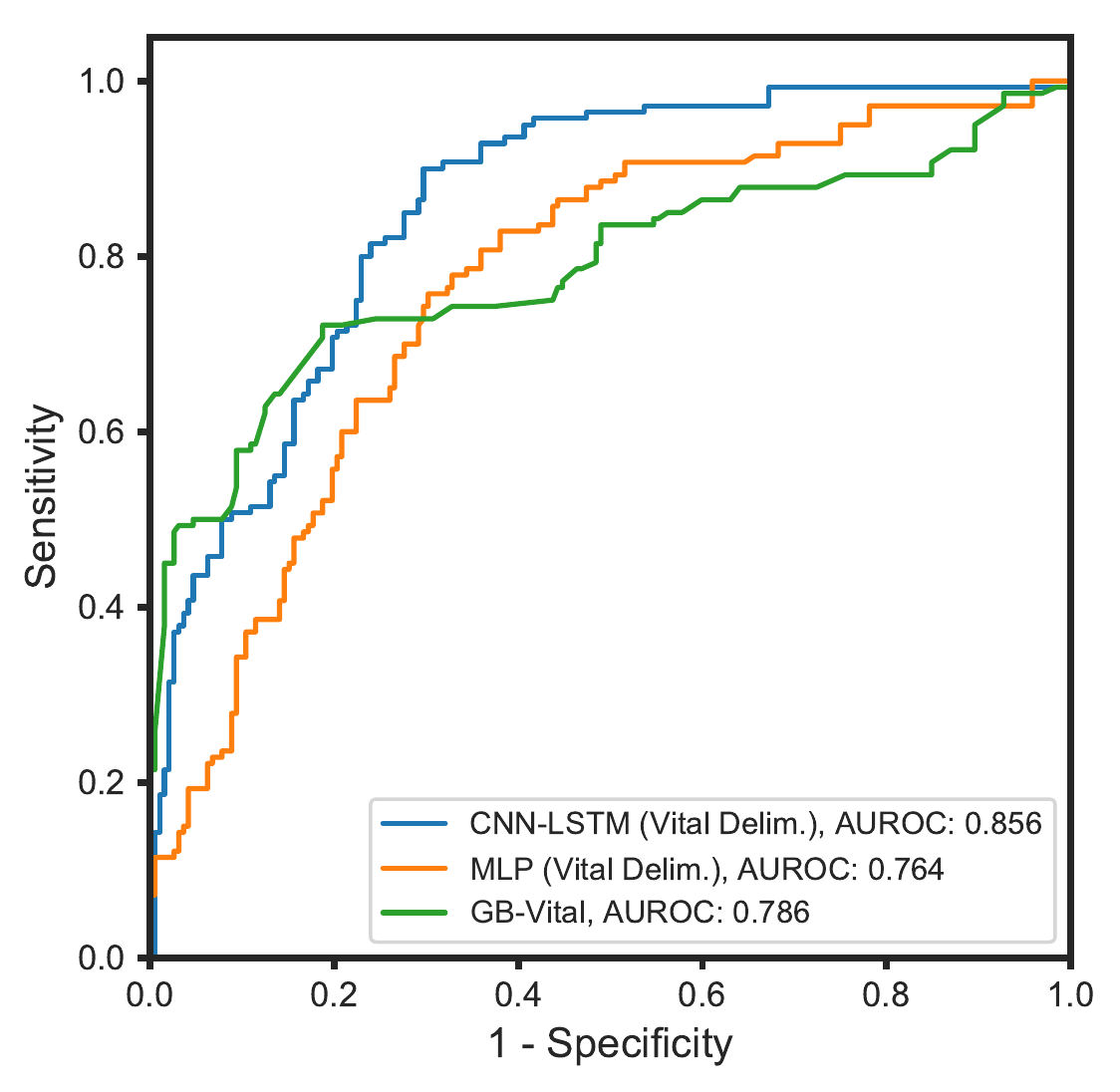}
                \caption{}
                \label{fig:5a}
        \end{subfigure}~%
        \begin{subfigure}[b]{0.336\textwidth}
                \includegraphics[width=\textwidth]{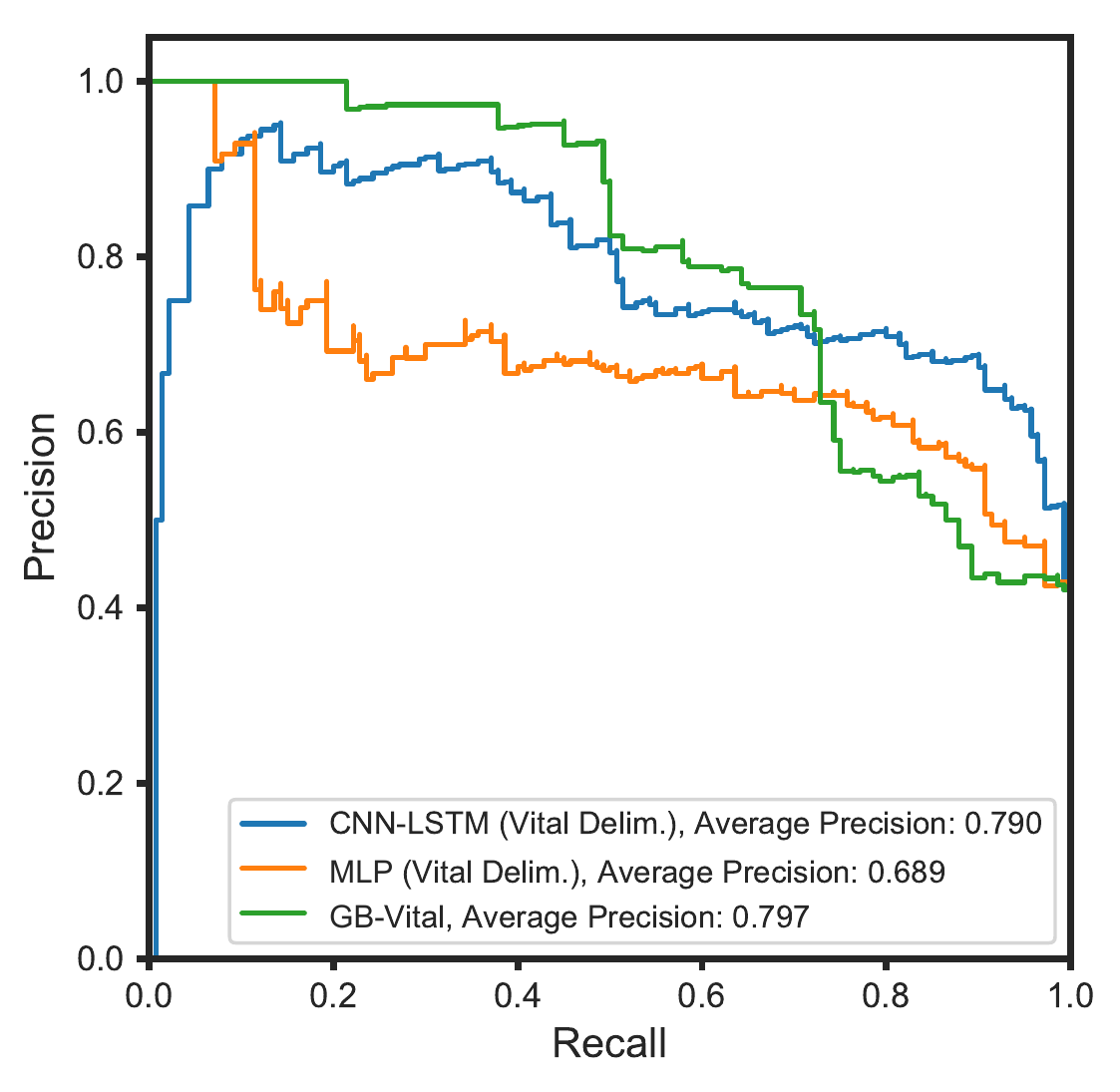}
                \caption{}
                \label{fig:5b}
        \end{subfigure}~%
        \begin{subfigure}[b]{0.336\textwidth}
                \includegraphics[width=\textwidth]{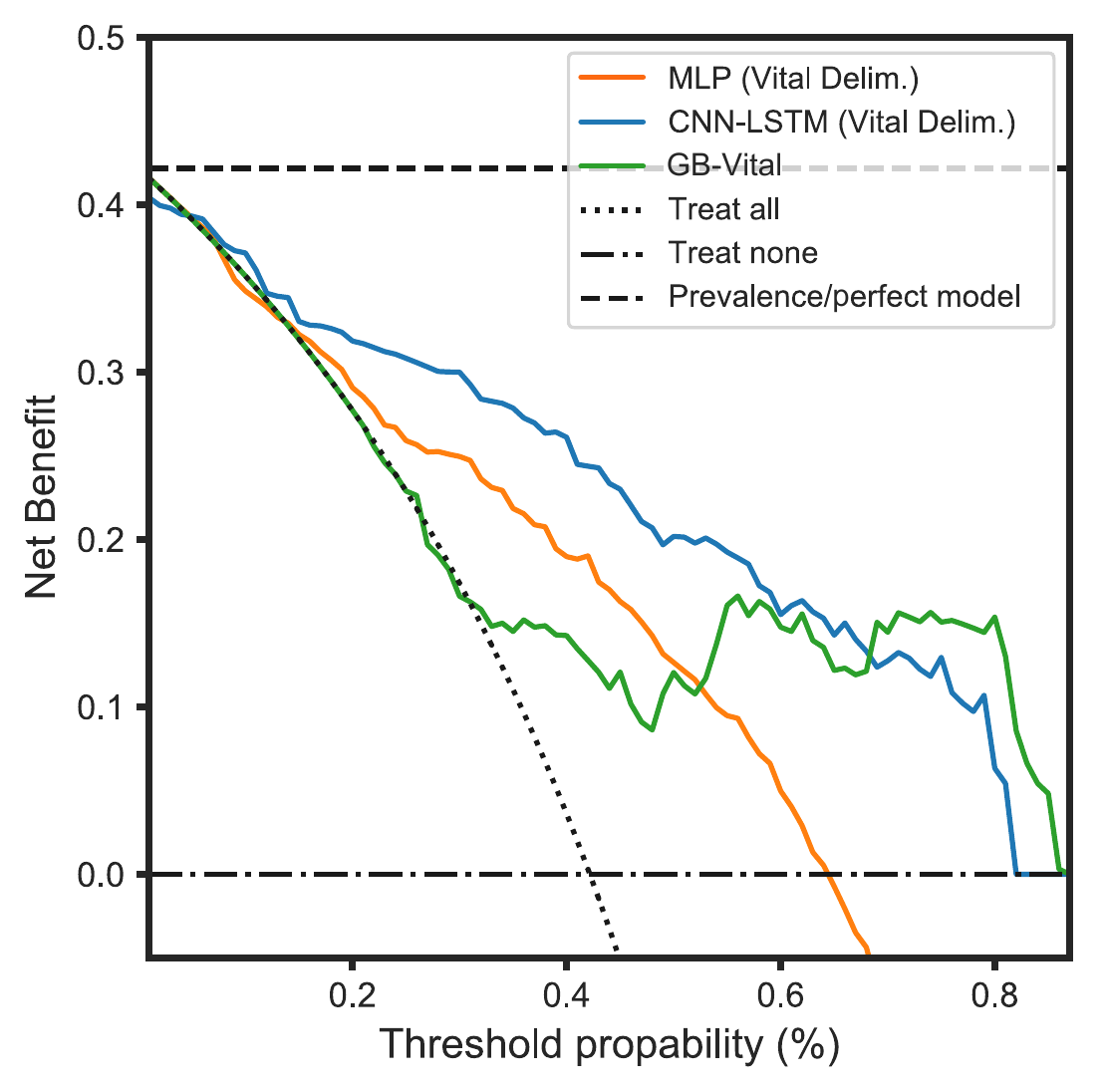}
                \caption{}
                \label{fig:5c}
        \end{subfigure}%
   		\setlength{\belowcaptionskip}{-12pt}
        \caption{Results from the vital sign test data set evaluated three hours before sepsis onset: a) ROC curves; b) PR curves; c) DCA.}
\end{figure*}

In Figure \ref{fig:5a} and \ref{fig:5b}, the ROC and PR curves from the vital sign test data set data are shown, respectively. The GB-Vital model achieved an AUROC of 0.786 and a mAP (mean average precision) of 0.797 when evaluated three hours before sepsis. Figure \ref{fig:5c} shows the results from the DCA. The NB of using the GB-Vital model was equal to the NB of treating all patients in the range of probability thresholds from 0$\%$  to 32$\%$. At thresholds above 32$\%$, the NB of using the GB-Vital model exceeded both the NB of treating none patients and the NB of treating all patients.
\par

\subsection{Multilayer perceptron}

The MLP model achieved an AUROC of 0.764 and a mAP of 0.689 when evaluated three hours before sepsis in the vital sign data set (Figure \ref{fig:5a} and \ref{fig:5b}). The NB of using the MLP model was equal to the NB of treating all patients in the range of probability thresholds from 0$\%$  to 20$\%$. At threshold values above 45$\%$, the NB of using the MLP model exceeded both the NB of treating no patients and the NB of treating all patients.
\par

\begin{figure*}[!h]
        \begin{subfigure}[b]{0.25\textwidth}
                \includegraphics[width=\textwidth]{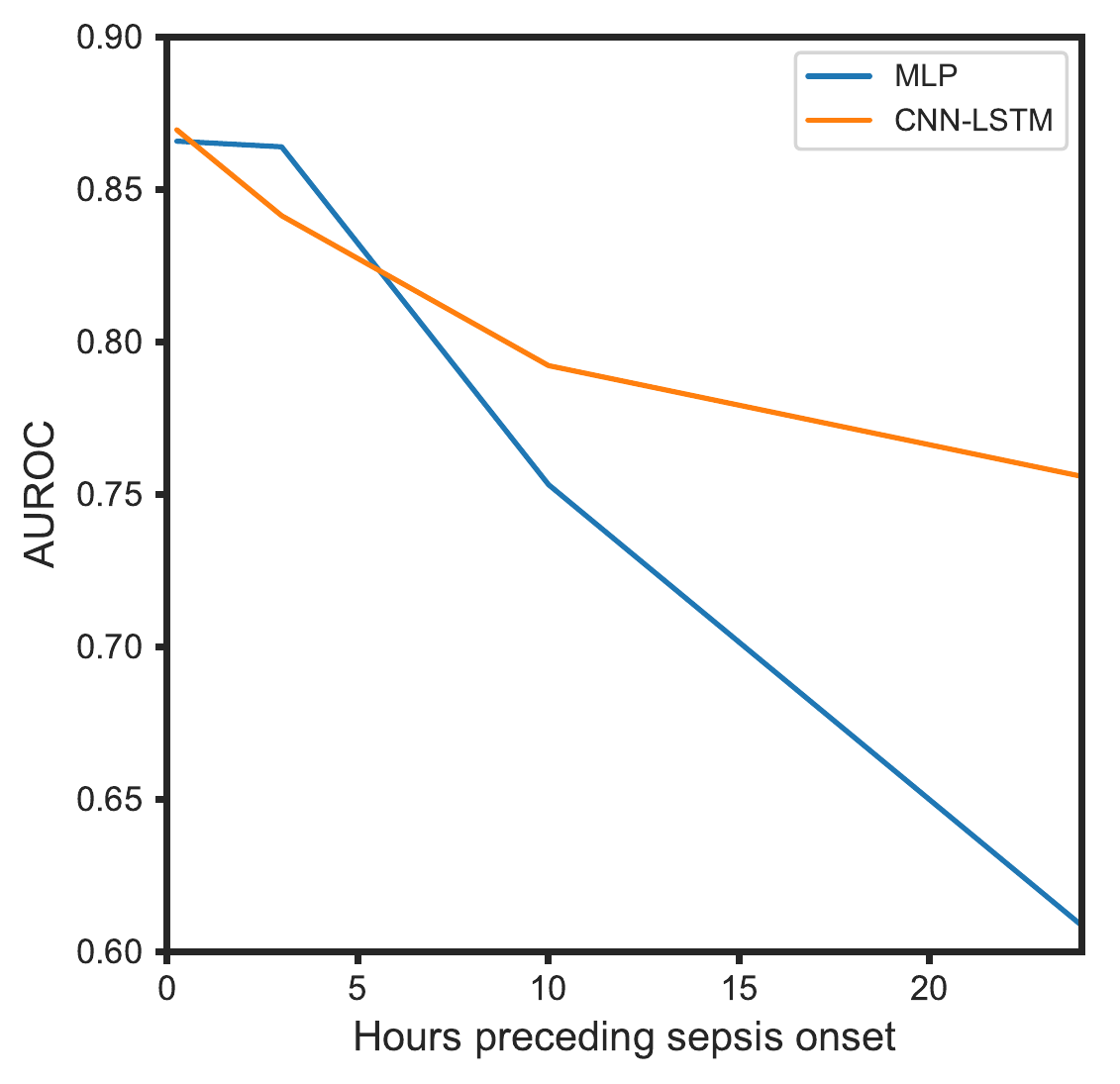}
                \caption{}
                \label{fig:6a}
        \end{subfigure}~%
        \begin{subfigure}[b]{0.25\textwidth}
                \includegraphics[width=\textwidth]{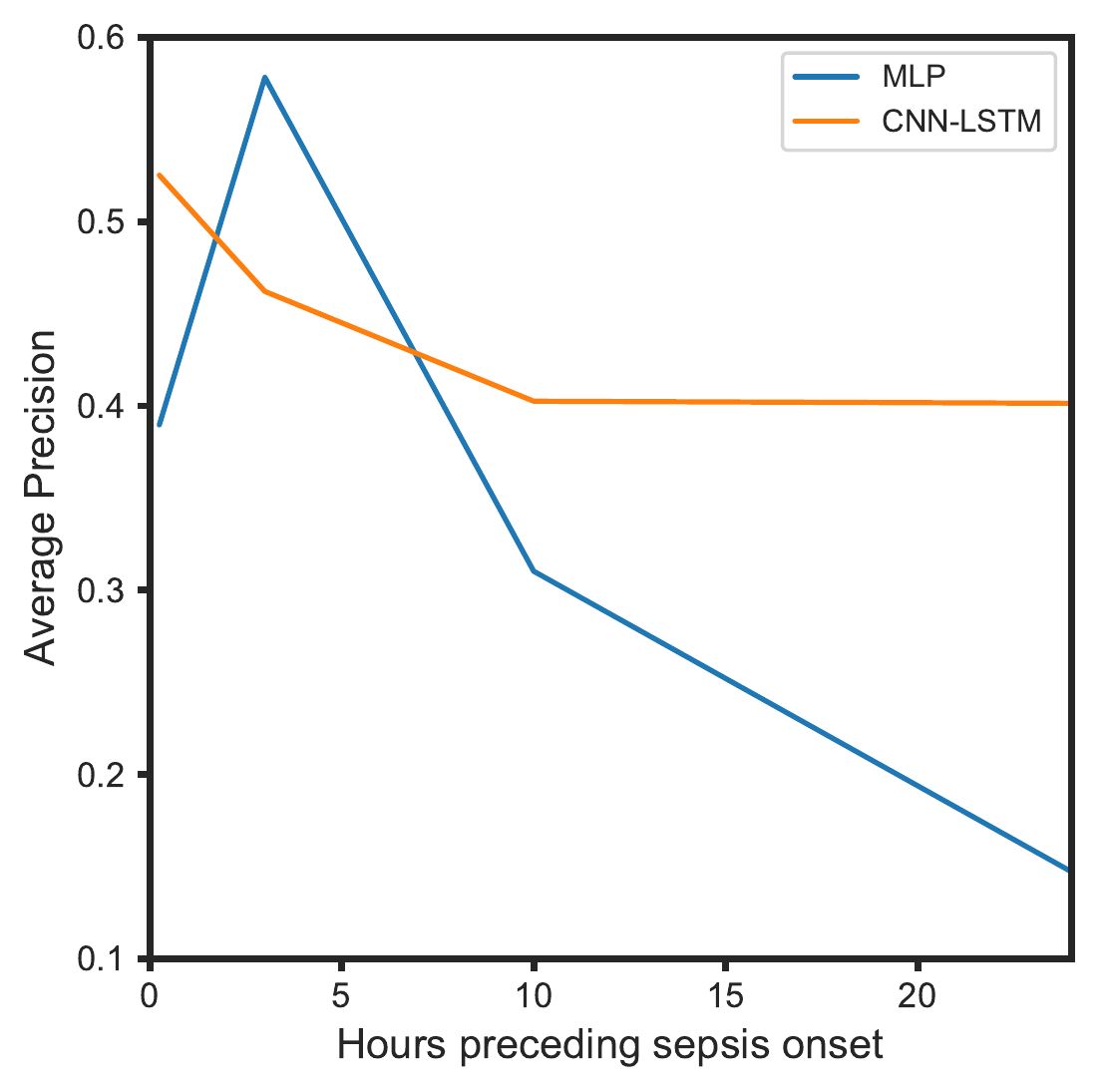}
                \caption{}
                \label{fig:6b}
        \end{subfigure}~%
        \begin{subfigure}[b]{0.25\textwidth}
                \includegraphics[width=\textwidth]{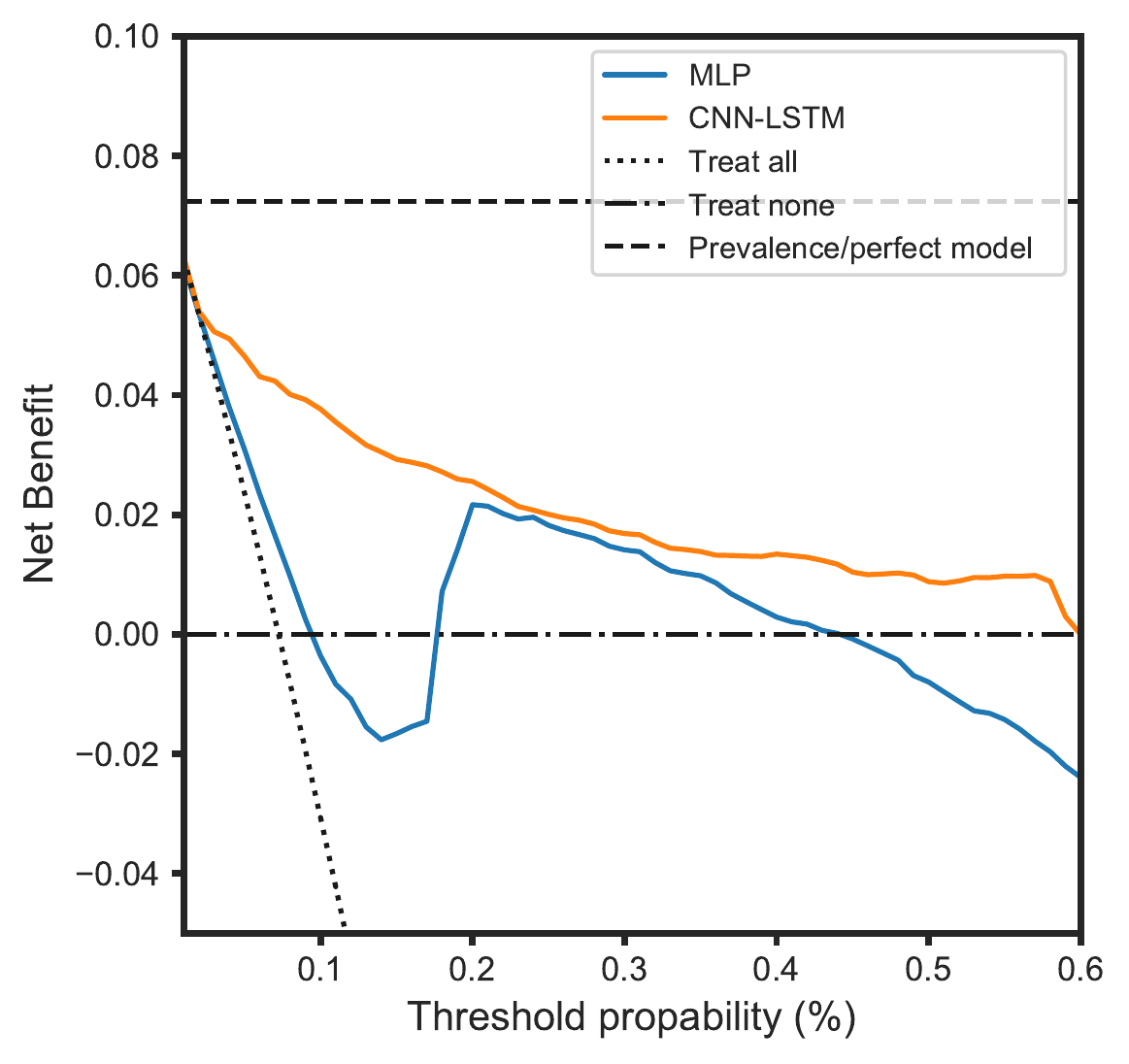}
                \caption{}
                \label{fig:6c}
        \end{subfigure}~%
        \begin{subfigure}[b]{0.25\textwidth}
                \includegraphics[width=\textwidth]{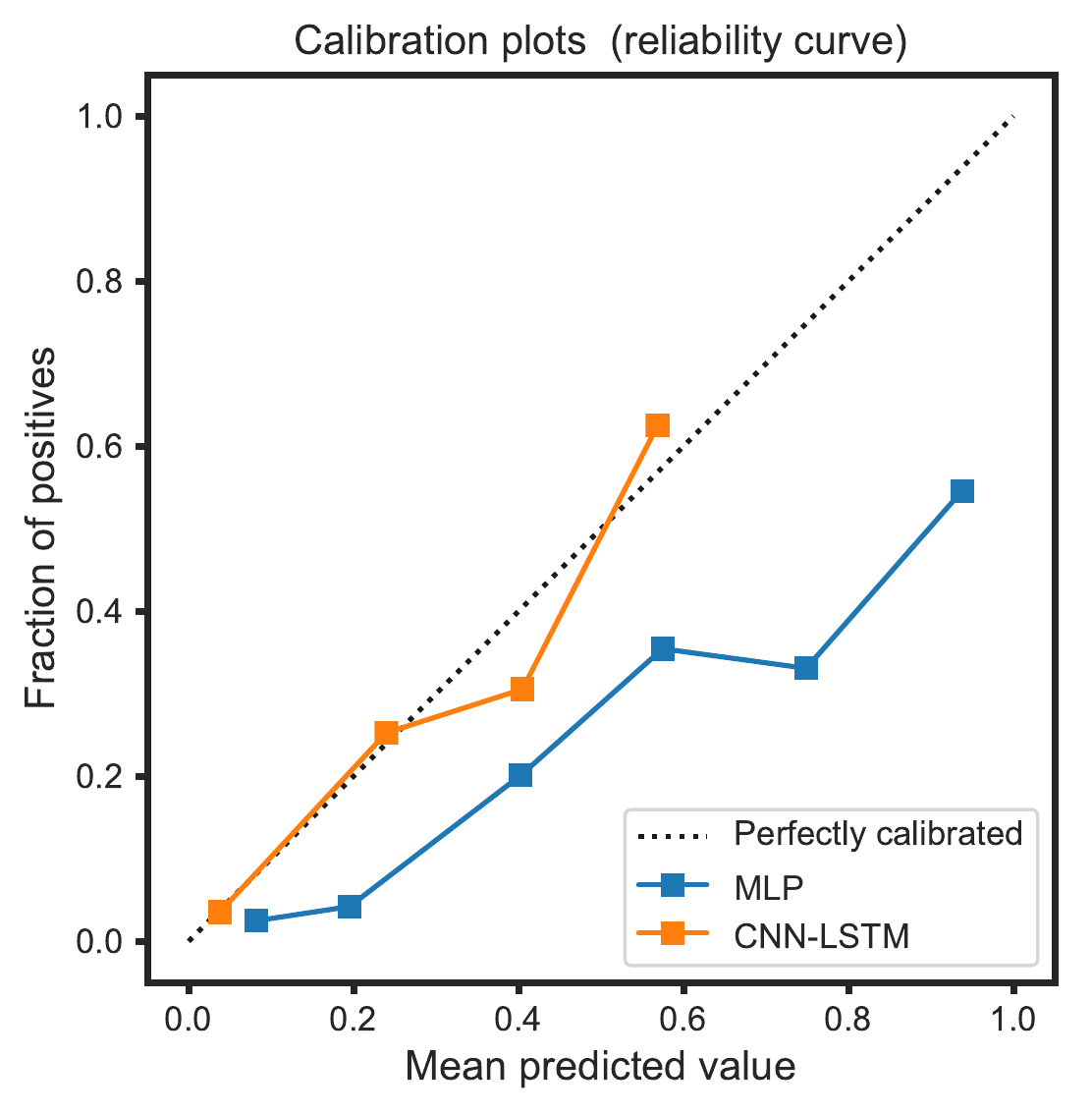}
                \caption{}
                \label{fig:6d}
        \end{subfigure}%
   		\setlength{\belowcaptionskip}{-12pt}

        \caption{Results from the full test data set: A) AUROC at different predictions times. B) PR curve at different prediction times. C) DCA three hours before sepsis onset. D) Calibration curve.}
        \label{fig:6}
\end{figure*}

Results from the full data set are summarized in Figure \ref{fig:6}. Figure \ref{fig:6a} and \ref{fig:6b} show how AUROC and mAP change as a function of time before the labeled onset of sepsis (or not sepsis). The MLP AUROC scores on the full data set were as follows: t-15 min: 0.872; t-3 hours:  0.871; t-10 hours: 0.751; and t-24 hours: 0.619. The highest mAP of 0.578 was achieved at three hours. The mAP dropped on both sides of this peak (0.395 at t-15 min and 0.318 at t-10 hours) and further decreased to 0.147 at t-24 hours.
\par

The NB of using the MLP model on the full data set was slightly higher than the NB of treating all patients in the range of probability thresholds from 0$\%$  to 12$\%$. In the range from 12$\%$  to 20$\%$  and above 45$\%$, the NB was negative. In the range from 20$\%$  to 45$\%$, the model exceeded both the NB of treating no patients and the NB of treating all patients (Figure \ref{fig:6c}).
\par

In Figure \ref{fig:6}, the calibration curve for the MLP model is shown (blue line). The plot provides an indication of whether future predicted probabilities agree with the observed probabilities. For example, if we predict a 35$\%$  risk of developing sepsis, the observed frequency of sepsis should be approximately 35 out of 100 patients with such a prediction. A perfectly calibrated model would have a 45 degree line along the diagonal \cite{Hilden_J_et_al_1978,Steyerberg_EW_et_al_2010}. 
\par

\subsection{CNN-LSTM}
The CNN-LSTM model achieved an AUROC of 0.856 and a mAP of 0.79 when evaluated three hours before sepsis on the vital sign test data (Figure \ref{fig:5a} and \ref{fig:5b}). Results from the full data set showed that the CNN-LSTM model decreased from a maximum mAP of 0.531 at t-15 min to 0.407 at both t-10 and t-24 hours. The CNN-LSTM AUROC scores were as follows: t-15 min: 0.879; t-3 hours: 0.842; t-10 hours: 0.792; and t-24 hours: 0.752. The NB of using the CNN-LSTM model on the full dataset exceeded both the NB of treating no patients and the NB of treating all patients in the range of probabilities from 5$\%$  to 60$\%$  (Figure \ref{fig:6c}). The calibration of the CNN-LSTM model is shown in Figure \ref{fig:6d} (orange line). 

\par

\subsection{SERAIP}
Table \ref{table2} shows the results of \textit{SERAIP.}. The columns $``$TP with IV antibiotics$"$, $``$TP with blood culture$"$, $``$TP with IV antibiotics or blood culture$"$  and $``$TP with no intervention$"$  indicate the potential for initiating interventions that the clinicians have not already thought about at the time of prediction. Using the first row of the table as an example, the model finds 17$\%$  of the positives, corresponding to 39 patients. The column $``$TP with IV antibiotics$"$  shows that 8 patients out of the 39 true positives were already started with intravenous antibiotics. The column $``$TP with blood culture$"$  shows that 12 out of the 39 patients had already had a blood culture, and $``$TP with IV antibiotics or blood culture$"$  shows that 8 of the 12 patients for whom intravenous antibiotics had been started had also had a blood culture. Finally, the last column $``$TP with no intervention$"$  shows that 27 of the 39 patients had no intervention initiated at the point of prediction.

The column "FP/FP" indicates the relationship between false positives and true positives and shows that one can expect 9.28 false alarms for every one true positive. 

\section{Discussion }
\subsection{Results}
We have presented an accurate deep learning system for early sepsis detection on a multi-center data set from outside ICUs. We have compared three different approaches for early detection of sepsis: a GB-Vital model, based on vital sign features; a non-sequential MLP model with thousands of features, including those used for the GB-Vital model; and a sequential CNN-LSTM model with an equal number of features. 
\par

\begin{table*}[!h]
\footnotesize
\adjustbox{max width=\linewidth}{%
\begin{tabular}{p{0.95in}p{0.6in}p{0.31in}p{0.36in}p{0.19in}p{0.06in}p{0.15in}p{0.1in}p{0.1in}p{0.35in}p{0.31in}p{0.31in}p{0.60in}}
\hline
\multicolumn{1}{p{0.95in}}{\textbf{Department/ Hospital}} & 
\multicolumn{1}{p{0.6in}}{\textbf{Evaluated up until}} & 
\multicolumn{1}{p{0.31in}}{\textbf{SEN}} & 
\multicolumn{1}{p{0.36in}}{\textbf{SPE}} & 
\multicolumn{1}{p{0.19in}}{ \textbf{FP/ TP}} & 
\multicolumn{1}{p{0.06in}}{ \textbf{TP}} & 
\multicolumn{1}{p{0.15in}}{\textbf{TN}} & 
\multicolumn{1}{p{0.1in}}{\textbf{FN}} & 
\multicolumn{1}{p{0.1in}}{\textbf{FP}} & 
\multicolumn{1}{p{0.35in}}{\textbf{TP anti}} & 
\multicolumn{1}{p{0.31in}}{\textbf{TP blood}} & 
\multicolumn{1}{p{0.31in}}{\textbf{TP int}} & 
\multicolumn{1}{p{0.60in}}{\textbf{TP no int.}} \\
\hline
\multicolumn{1}{p{0.95in}}{\multirow{3}{*}{\begin{tabular}{p{0.95in}}Emergency Dept. \newline
Hosp. 1\\\end{tabular}}} & 
\multicolumn{1}{p{0.6in}}{ t-3 hours} & 
\multicolumn{1}{p{0.31in}}{ 0.17} & 
\multicolumn{1}{p{0.36in}}{ 0.91} & 
\multicolumn{1}{p{0.19in}}{ 9.28} & 
\multicolumn{1}{p{0.06in}}{39} & 
\multicolumn{1}{p{0.15in}}{3663} & 
\multicolumn{1}{p{0.1in}}{ 197} & 
\multicolumn{1}{p{0.1in}}{ 362} & 
\multicolumn{1}{p{0.35in}}{ 8} & 
\multicolumn{1}{p{0.31in}}{ 12} & 
\multicolumn{1}{p{0.31in}}{ 12} & 
\multicolumn{1}{p{0.44in}}{ 27} \\
\hhline{~------------}
\multicolumn{1}{p{0.95in}}{} & 
\multicolumn{1}{p{0.6in}}{ t-10 hours} & 
\multicolumn{1}{p{0.31in}}{ 0.13} & 
\multicolumn{1}{p{0.36in}}{ 0.91} & 
\multicolumn{1}{p{0.19in}}{ 14.77} & 
\multicolumn{1}{p{0.06in}}{ 22} & 
\multicolumn{1}{p{0.15in}}{ 3456} & 
\multicolumn{1}{p{0.1in}}{ 151} & 
\multicolumn{1}{p{0.1in}}{ 325} & 
\multicolumn{1}{p{0.35in}}{ 4} & 
\multicolumn{1}{p{0.31in}}{ 7} & 
\multicolumn{1}{p{0.31in}}{ 7} & 
\multicolumn{1}{p{0.44in}}{ 15} \\
\hhline{~------------}
\multicolumn{1}{p{0.95in}}{} & 
\multicolumn{1}{p{0.6in}}{ t-24 hours} & 
\multicolumn{1}{p{0.31in}}{ 0.11} & 
\multicolumn{1}{p{0.36in}}{ 0.90} & 
\multicolumn{1}{p{0.19in}}{ 18.76} & 
\multicolumn{1}{p{0.06in}}{ 17} & 
\multicolumn{1}{p{0.15in}}{ 2899} & 
\multicolumn{1}{p{0.1in}}{ 132} & 
\multicolumn{1}{p{0.1in}}{ 319} & 
\multicolumn{1}{p{0.35in}}{ 1} & 
\multicolumn{1}{p{0.31in}}{ 2} & 
\multicolumn{1}{p{0.31in}}{ 2} & 
\multicolumn{1}{p{0.44in}}{ 15} \\
\hline
\multicolumn{1}{p{0.95in}}{\multirow{3}{*}{\begin{tabular}{p{0.95in}}Dept. of \newline Oncology\newline
Hosp. 2\\\end{tabular}}} & 
\multicolumn{1}{p{0.6in}}{ t-3 hours} & 
\multicolumn{1}{p{0.31in}}{ 0.31} & 
\multicolumn{1}{p{0.36in}}{ 0.93} & 
\multicolumn{1}{p{0.19in}}{ 7.25} & 
\multicolumn{1}{p{0.06in}}{ 4} & 
\multicolumn{1}{p{0.15in}}{ 389} & 
\multicolumn{1}{p{0.1in}}{ 9} & 
\multicolumn{1}{p{0.1in}}{ 29} & 
\multicolumn{1}{p{0.35in}}{ 3} & 
\multicolumn{1}{p{0.31in}}{ 1} & 
\multicolumn{1}{p{0.31in}}{ 3} & 
\multicolumn{1}{p{0.44in}}{ 1} \\
\hhline{~------------}
\multicolumn{1}{p{0.8in}}{} & 
\multicolumn{1}{p{0.6in}}{ t-10 hours} & 
\multicolumn{1}{p{0.31in}}{ 0.40} & 
\multicolumn{1}{p{0.36in}}{ 0.93} & 
\multicolumn{1}{p{0.19in}}{ 7.25} & 
\multicolumn{1}{p{0.06in}}{ 4} & 
\multicolumn{1}{p{0.15in}}{ 372} & 
\multicolumn{1}{p{0.1in}}{ 6} & 
\multicolumn{1}{p{0.1in}}{ 29} & 
\multicolumn{1}{p{0.35in}}{ 2} & 
\multicolumn{1}{p{0.31in}}{ 1} & 
\multicolumn{1}{p{0.31in}}{ 3} & 
\multicolumn{1}{p{0.44in}}{ 1} \\
\hhline{~------------}
\multicolumn{1}{p{0.8in}}{} & 
\multicolumn{1}{p{0.6in}}{ t-24 hours} & 
\multicolumn{1}{p{0.31in}}{ 0.30} & 
\multicolumn{1}{p{0.36in}}{ 0.93} & 
\multicolumn{1}{p{0.19in}}{ 9.33} & 
\multicolumn{1}{p{0.06in}}{ 3} & 
\multicolumn{1}{p{0.15in}}{ 364} & 
\multicolumn{1}{p{0.1in}}{ 7} & 
\multicolumn{1}{p{0.1in}}{ 28} & 
\multicolumn{1}{p{0.35in}}{ 2} & 
\multicolumn{1}{p{0.31in}}{ 0} & 
\multicolumn{1}{p{0.31in}}{ 2} & 
\multicolumn{1}{p{0.44in}}{ 1} \\
\hline
\multicolumn{1}{p{0.8in}}{\multirow{3}{*}{\begin{tabular}{p{0.95in}}Joint \newline Emergency
Dept. \newline
Hosp. 1\\\end{tabular}}} & 
\multicolumn{1}{p{0.6in}}{ t-3 hours} & 
\multicolumn{1}{p{0.31in}}{ 0.33} & 
\multicolumn{1}{p{0.36in}}{ 0.87} & 
\multicolumn{1}{p{0.19in}}{ 3.00} & 
\multicolumn{1}{p{0.06in}}{ 4} & 
\multicolumn{1}{p{0.15in}}{ 78} & 
\multicolumn{1}{p{0.1in}}{ 8} & 
\multicolumn{1}{p{0.1in}}{ 12} & 
\multicolumn{1}{p{0.35in}}{ 0} & 
\multicolumn{1}{p{0.31in}}{ 1} & 
\multicolumn{1}{p{0.31in}}{ 1} & 
\multicolumn{1}{p{0.44in}}{ 3} \\
\hhline{~------------}
\multicolumn{1}{p{0.8in}}{} & 
\multicolumn{1}{p{0.6in}}{ t-10 hours} & 
\multicolumn{1}{p{0.31in}}{ 0.09} & 
\multicolumn{1}{p{0.36in}}{ 0.10} & 
\multicolumn{1}{p{0.19in}}{ 1.50} & 
\multicolumn{1}{p{0.06in}}{ 6} & 
\multicolumn{1}{p{0.15in}}{ 1} & 
\multicolumn{1}{p{0.1in}}{ 60} & 
\multicolumn{1}{p{0.1in}}{ 9} & 
\multicolumn{1}{p{0.35in}}{ 0} & 
\multicolumn{1}{p{0.31in}}{ 1} & 
\multicolumn{1}{p{0.31in}}{ 1} & 
\multicolumn{1}{p{0.44in}}{ 5} \\
\hhline{~------------}
\multicolumn{1}{p{0.8in}}{} & 
\multicolumn{1}{p{0.6in}}{ t-24 hours} & 
\multicolumn{1}{p{0.31in}}{ 0.09} & 
\multicolumn{1}{p{0.36in}}{ 0.14} & 
\multicolumn{1}{p{0.19in}}{ 1.20} & 
\multicolumn{1}{p{0.06in}}{ 5} & 
\multicolumn{1}{p{0.15in}}{ 1} & 
\multicolumn{1}{p{0.1in}}{ 50} & 
\multicolumn{1}{p{0.1in}}{ 6} & 
\multicolumn{1}{p{0.35in}}{ 0} & 
\multicolumn{1}{p{0.31in}}{ 0} & 
\multicolumn{1}{p{0.31in}}{ 0} & 
\multicolumn{1}{p{0.44in}}{ 5} \\
\hline
\multicolumn{1}{p{0.95in}}{\multirow{3}{*}{\begin{tabular}{p{0.95in}}Emergency Dept. \newline
Hosp. 3\\\end{tabular}}} & 
\multicolumn{1}{p{0.6in}}{ t-3 hours} & 
\multicolumn{1}{p{0.31in}}{ 1.00} & 
\multicolumn{1}{p{0.36in}}{ 0.92} & 
\multicolumn{1}{p{0.19in}}{ 5.00} & 
\multicolumn{1}{p{0.06in}}{ 1} & 
\multicolumn{1}{p{0.15in}}{ 55} & 
\multicolumn{1}{p{0.1in}}{ 0} & 
\multicolumn{1}{p{0.1in}}{ 5} & 
\multicolumn{1}{p{0.35in}}{ 0} & 
\multicolumn{1}{p{0.31in}}{ 0} & 
\multicolumn{1}{p{0.31in}}{ 0} & 
\multicolumn{1}{p{0.44in}}{ 1} \\
\hhline{~------------}
\multicolumn{1}{p{0.8in}}{} & 
\multicolumn{1}{p{0.6in}}{ t-10 hours} & 
\multicolumn{1}{p{0.31in}}{ 1.00} & 
\multicolumn{1}{p{0.36in}}{ 0.86} & 
\multicolumn{1}{p{0.19in}}{ 5.00} & 
\multicolumn{1}{p{0.06in}}{ 1} & 
\multicolumn{1}{p{0.15in}}{ 31} & 
\multicolumn{1}{p{0.1in}}{ 0} & 
\multicolumn{1}{p{0.1in}}{ 5} & 
\multicolumn{1}{p{0.35in}}{ 0} & 
\multicolumn{1}{p{0.31in}}{ 0} & 
\multicolumn{1}{p{0.31in}}{ 0} & 
\multicolumn{1}{p{0.44in}}{ 1} \\
\hhline{~------------}
\multicolumn{1}{p{0.8in}}{} & 
\multicolumn{1}{p{0.6in}}{ t-24 hours} & 
\multicolumn{1}{p{0.31in}}{ 1.00} & 
\multicolumn{1}{p{0.36in}}{ 0.85} & 
\multicolumn{1}{p{0.19in}}{ 5.00} & 
\multicolumn{1}{p{0.06in}}{ 1} & 
\multicolumn{1}{p{0.15in}}{ 28} & 
\multicolumn{1}{p{0.1in}}{ 0} & 
\multicolumn{1}{p{0.1in}}{ 5} & 
\multicolumn{1}{p{0.35in}}{ 0} & 
\multicolumn{1}{p{0.31in}}{ 0} & 
\multicolumn{1}{p{0.31in}}{ 0} & 
\multicolumn{1}{p{0.44in}}{ 1} \\
\hline
\multicolumn{1}{p{0.8in}}{\multirow{3}{*}{\begin{tabular}{p{0.95in}}Dept. of \newline
Anaesthesiology \newline
Hosp. 1\\\end{tabular}}} & 
\multicolumn{1}{p{0.6in}}{ t-3 hours} & 
\multicolumn{1}{p{0.31in}}{ 0.60} & 
\multicolumn{1}{p{0.36in}}{ 0.66} & 
\multicolumn{1}{p{0.19in}}{ 1.83} & 
\multicolumn{1}{p{0.06in}}{ 6} & 
\multicolumn{1}{p{0.15in}}{ 21} & 
\multicolumn{1}{p{0.1in}}{ 4} & 
\multicolumn{1}{p{0.1in}}{ 11} & 
\multicolumn{1}{p{0.35in}}{ 3} & 
\multicolumn{1}{p{0.31in}}{ 2} & 
\multicolumn{1}{p{0.31in}}{ 3} & 
\multicolumn{1}{p{0.44in}}{ 3} \\
\hhline{~------------}
\multicolumn{1}{p{0.8in}}{} & 
\multicolumn{1}{p{0.6in}}{ t-10 hours} & 
\multicolumn{1}{p{0.31in}}{ 0.45} & 
\multicolumn{1}{p{0.36in}}{ 0.59} & 
\multicolumn{1}{p{0.19in}}{ 2.20} & 
\multicolumn{1}{p{0.06in}}{ 5} & 
\multicolumn{1}{p{0.15in}}{ 16} & 
\multicolumn{1}{p{0.1in}}{ 6} & 
\multicolumn{1}{p{0.1in}}{ 11} & 
\multicolumn{1}{p{0.35in}}{ 1} & 
\multicolumn{1}{p{0.31in}}{ 1} & 
\multicolumn{1}{p{0.31in}}{ 1} & 
\multicolumn{1}{p{0.44in}}{ 4} \\
\hhline{~------------}
\multicolumn{1}{p{0.8in}}{} & 
\multicolumn{1}{p{0.6in}}{ t-24 hours} & 
\multicolumn{1}{p{0.31in}}{ 0.56} & 
\multicolumn{1}{p{0.36in}}{ 0.60} & 
\multicolumn{1}{p{0.19in}}{ 2.00} & 
\multicolumn{1}{p{0.06in}}{ 5} & 
\multicolumn{1}{p{0.15in}}{ 15} & 
\multicolumn{1}{p{0.1in}}{ 4} & 
\multicolumn{1}{p{0.1in}}{ 10} & 
\multicolumn{1}{p{0.35in}}{ 0} & 
\multicolumn{1}{p{0.31in}}{ 1} & 
\multicolumn{1}{p{0.31in}}{ 1} & 
\multicolumn{1}{p{0.44in}}{ 4} \\
\hline
\multicolumn{1}{p{0.95in}}{\multirow{3}{*}{\begin{tabular}{p{0.95in}}Dept. of \newline Hematology \newline
Hosp. 2\\\end{tabular}}} & 
\multicolumn{1}{p{0.6in}}{ t-3 hours} & 
\multicolumn{1}{p{0.31in}}{ 0.36} & 
\multicolumn{1}{p{0.36in}}{ 0.93} & 
\multicolumn{1}{p{0.19in}}{ 5.80} & 
\multicolumn{1}{p{0.06in}}{ 5} & 
\multicolumn{1}{p{0.15in}}{ 398} & 
\multicolumn{1}{p{0.1in}}{ 9} & 
\multicolumn{1}{p{0.1in}}{ 29} & 
\multicolumn{1}{p{0.35in}}{ 3} & 
\multicolumn{1}{p{0.31in}}{ 1} & 
\multicolumn{1}{p{0.31in}}{ 3} & 
\multicolumn{1}{p{0.44in}}{ 2} \\
\hhline{~------------}
\multicolumn{1}{p{0.8in}}{} & 
\multicolumn{1}{p{0.6in}}{ t-10 hours} & 
\multicolumn{1}{p{0.31in}}{ 0.45} & 
\multicolumn{1}{p{0.36in}}{ 0.93} & 
\multicolumn{1}{p{0.19in}}{ 5.80} & 
\multicolumn{1}{p{0.06in}}{ 5} & 
\multicolumn{1}{p{0.15in}}{ 372} & 
\multicolumn{1}{p{0.1in}}{ 6} & 
\multicolumn{1}{p{0.1in}}{ 29} & 
\multicolumn{1}{p{0.35in}}{ 2} & 
\multicolumn{1}{p{0.31in}}{ 1} & 
\multicolumn{1}{p{0.31in}}{ 3} & 
\multicolumn{1}{p{0.44in}}{ 2} \\
\hhline{~------------}
\multicolumn{1}{p{0.8in}}{} & 
\multicolumn{1}{p{0.6in}}{ t-24 hours} & 
\multicolumn{1}{p{0.31in}}{ 0.30} & 
\multicolumn{1}{p{0.36in}}{ 0.93} & 
\multicolumn{1}{p{0.19in}}{ 9.33} & 
\multicolumn{1}{p{0.06in}}{ 3} & 
\multicolumn{1}{p{0.15in}}{ 364} & 
\multicolumn{1}{p{0.1in}}{ 7} & 
\multicolumn{1}{p{0.1in}}{ 28} & 
\multicolumn{1}{p{0.35in}}{ 2} & 
\multicolumn{1}{p{0.31in}}{ 0} & 
\multicolumn{1}{p{0.31in}}{ 2} & 
\multicolumn{1}{p{0.44in}}{ 1} \\
\hline
\multicolumn{1}{p{0.95in}}{\multirow{3}{*}{\begin{tabular}{p{0.95in}}Dept. of \newline
gastroin. surgery \newline
Hosp. 2\\\end{tabular}}} & 
\multicolumn{1}{p{0.6in}}{ t-3 hours} & 
\multicolumn{1}{p{0.31in}}{ 0.67} & 
\multicolumn{1}{p{0.36in}}{ 0.63} & 
\multicolumn{1}{p{0.19in}}{ 2.75} & 
\multicolumn{1}{p{0.06in}}{ 4} & 
\multicolumn{1}{p{0.15in}}{ 19} & 
\multicolumn{1}{p{0.1in}}{ 2} & 
\multicolumn{1}{p{0.1in}}{ 11} & 
\multicolumn{1}{p{0.35in}}{ 1} & 
\multicolumn{1}{p{0.31in}}{ 0} & 
\multicolumn{1}{p{0.31in}}{ 1} & 
\multicolumn{1}{p{0.44in}}{ 3} \\
\hhline{~------------}
\multicolumn{1}{p{0.8in}}{} & 
\multicolumn{1}{p{0.6in}}{ t-10 hours} & 
\multicolumn{1}{p{0.31in}}{ 1.00} & 
\multicolumn{1}{p{0.36in}}{ 0.62} & 
\multicolumn{1}{p{0.19in}}{ 2.75} & 
\multicolumn{1}{p{0.06in}}{ 4} & 
\multicolumn{1}{p{0.15in}}{ 18} & 
\multicolumn{1}{p{0.1in}}{ 0} & 
\multicolumn{1}{p{0.1in}}{ 11} & 
\multicolumn{1}{p{0.35in}}{ 1} & 
\multicolumn{1}{p{0.31in}}{ 0} & 
\multicolumn{1}{p{0.31in}}{ 1} & 
\multicolumn{1}{p{0.44in}}{ 3} \\
\hhline{~------------}
\multicolumn{1}{p{0.8in}}{} & 
\multicolumn{1}{p{0.6in}}{ t-24 hours} & 
\multicolumn{1}{p{0.31in}}{ 1.00} & 
\multicolumn{1}{p{0.36in}}{ 0.63} & 
\multicolumn{1}{p{0.19in}}{ 2.50} & 
\multicolumn{1}{p{0.06in}}{ 4} & 
\multicolumn{1}{p{0.15in}}{ 17} & 
\multicolumn{1}{p{0.1in}}{ 0} & 
\multicolumn{1}{p{0.1in}}{ 10} & 
\multicolumn{1}{p{0.35in}}{ 1} & 
\multicolumn{1}{p{0.31in}}{ 0} & 
\multicolumn{1}{p{0.31in}}{ 1} & 
\multicolumn{1}{p{0.44in}}{ 3} \\
\hline
\multicolumn{1}{p{0.95in}}{\multirow{3}{*}{\begin{tabular}{p{0.95in}}Dept. of \newline
Anaesthesiology \newline
Hosp. 2\\\end{tabular}}} & 
\multicolumn{1}{p{0.6in}}{ t-3 hours} & 
\multicolumn{1}{p{0.31in}}{ 0.33} & 
\multicolumn{1}{p{0.36in}}{ 0.95} & 
\multicolumn{1}{p{0.19in}}{ 1.00} & 
\multicolumn{1}{p{0.06in}}{ 1} & 
\multicolumn{1}{p{0.15in}}{ 19} & 
\multicolumn{1}{p{0.1in}}{ 2} & 
\multicolumn{1}{p{0.1in}}{ 1} & 
\multicolumn{1}{p{0.35in}}{ 0} & 
\multicolumn{1}{p{0.31in}}{ 0} & 
\multicolumn{1}{p{0.31in}}{ 0} & 
\multicolumn{1}{p{0.44in}}{ 1} \\
\hhline{~------------}
\multicolumn{1}{p{0.8in}}{} & 
\multicolumn{1}{p{0.6in}}{ t-10 hours} & 
\multicolumn{1}{p{0.31in}}{ 0.50} & 
\multicolumn{1}{p{0.36in}}{ 0.95} & 
\multicolumn{1}{p{0.19in}}{ 1.00} & 
\multicolumn{1}{p{0.06in}}{ 1} & 
\multicolumn{1}{p{0.15in}}{ 19} & 
\multicolumn{1}{p{0.1in}}{ 1} & 
\multicolumn{1}{p{0.1in}}{ 1} & 
\multicolumn{1}{p{0.35in}}{ 0} & 
\multicolumn{1}{p{0.31in}}{ 0} & 
\multicolumn{1}{p{0.31in}}{ 0} & 
\multicolumn{1}{p{0.44in}}{ 1} \\
\hhline{~------------}
\multicolumn{1}{p{0.8in}}{} & 
\multicolumn{1}{p{0.6in}}{ t-24 hours} & 
\multicolumn{1}{p{0.31in}}{ 0.50} & 
\multicolumn{1}{p{0.36in}}{ 0.95} & 
\multicolumn{1}{p{0.19in}}{ 1.00} & 
\multicolumn{1}{p{0.06in}}{ 1} & 
\multicolumn{1}{p{0.15in}}{ 18} & 
\multicolumn{1}{p{0.1in}}{ 1} & 
\multicolumn{1}{p{0.1in}}{ 1} & 
\multicolumn{1}{p{0.35in}}{ 0} & 
\multicolumn{1}{p{0.31in}}{ 0} & 
\multicolumn{1}{p{0.31in}}{ 0} & 
\multicolumn{1}{p{0.44in}}{ 1} \\
\hline
\end{tabular}}
\caption{Results from the ``sequence evaluation with retrospective assessment of intervention potential'' on the full test data set.
Area under the receiver operating characteristics (AUROC), false positive (FP), true negative (TN), false negative (FN), false positive
(FP), intravenous (IV), Hospital (Hosp.), gastrointestinal (gastroin.), specificity (SPE) sensitivity (SEN), TP with IV antibiotics (TP anti), TP with blood culture (TP blood), TP with IV antibiotics or blood culture (TP int.), TP with no intervention (TP no int.).}
\label{table2}
\end{table*}


The GB-Vital model had reasonable performance, with an AUROC of 0.786 three hours before sepsis onset for patients with registered vital signs, but it underperforms when compared to the results of previous studies on early sepsis detection. Qingqing Mao et al. achieved an AUROC of 0.88 \cite{Qingqing_Mao_et_al_2018} three hours before sepsis onset with a similar GB-Vital model. Nemati et al. reported an AUROC of 0.85 four hours before sepsis onset with a Weilbull--Cox proportional hazards model. Futoma et al. reported AUROCs of 0.86 and 0.78 three and twelve hours before sepsis onset, respectively, with a multi-output Gaussian processes model \cite{Joseph_Futoma_et_al_2017AI}. 
The reason for the lower AUROC values in our GB-Vital model is likely due to the amount of missing values in our data set. All of the above studies are built solely on data from ICUs, where vital parameters are recorded frequently. Recall that in our diverse data set, only 65$\%$ of the sepsis patients had at least two vital signs measured three hours before sepsis onset. Qingqing Mao et al. examined the direct impact of missing values in their GB-Vital model and found that AUROC decreased from 0.9 to 0.79 when increasing the percentage of missing values from 0$\%$  to 20$\%$. Increasing the percentage of missing values even further to 60$\%$  yielded an AUROC of 0.75. The reported AUROC of 0.79 with a missing value rate of 20$\%$ is directly comparable to our GB-Vital model, which archived an AUROC of 0.786 three hours before sepsis with a similar rate of missing values. It is important to note that when Qingqing Mao et al. created a data set reminiscent of ours, our results correlate.

These numbers indicate that although the GB-Vital model performs well on ICU data, it may not be useful for the early detection of sepsis at a broader scale, where vital parameters are not recorded as frequently across various hospital departments.
\par

In Figure \ref{fig:6a}, it can be seen that the MLP and CNN-LSTM models had close to equal AUROC performance at time t-3, and in fact, the MLP model had a better average precision (Figure \ref{fig:6b}) than the CNN-LSTM model. This was probably because the DNN model was trained on data three hours before sepsis onset. In contrast, the CNN-LSTM model appeared to be more stable when used at different times relative to sepsis onset, which may be attributed to the sequential modeling approach. 
\par

The CNN-LSTM had higher NB values in the DCA compared to the MLP model for the full range of threshold values (Figure \ref{fig:6c}). In addition, a slightly odd NB profile could be observed for the MLP model in the threshold range from 0.05 to 0.2, indicating that the model was not well calibrated in this area and therefore would serve poorly as a risk-estimation model. This was investigated with a calibration plot, as shown in Figure \ref{fig:6d}. The plot supported our presumption that the MLP model was poorly calibrated, as the observed frequency of sepsis was systematically higher than the predicted risk of developing sepsis, especially in the probability ranges of 0.05--0.2 and 0.6--0.8. The CNN-LSTM model did not seem to suffer from poor calibration.
\par

\subsection{SERAIP}
\textit{The SERAIP} is our attempt to create an $``$close to the clinic$"$  evaluation yielding an accurate picture of how the algorithm could support the clinical work at different departments. SERAIP can be considered an extension of the real-time validation suggested by Futoma et al. \cite{Joseph_Futoma_et_al_2017AI}.

We simulated real-world usage by doing a retrospective evaluation, investigating two of the most important actions that a sepsis detection model could help initiate. Intravenous antibiotics and blood culture requisitions has been analyzed in the period preceding predictions, allowing for better estimates of the clinical utility of the model. The numbers for sensitivity and specificity in Table \ref{table2} were calculated using a global probability threshold of 0.1, which was determined from inspection of ROC and DCA. Optimally, a threshold should have been chosen per department, as the patient case mix varies greatly. 

Looking at the Emergency department, Hospital 1 the model had a sensitivity of 0.17 three hours before sepsis, which corresponded to finding 39 true positives, of which 27 had not received any intervention. Conversely, as many as 362 false positives (at a specificity of 0.91) must be accepted at a threshold of 0.1. The hematology department (Hospital 2) is an example of a department with a completely different patient clientele than the emergency department. Here the model had a sensitivity of 0.45 and a specificity of 0.93 ten hours before sepsis onset, which corresponded to 5 true positives and 29 false positives. At the same time, no interventions were initiated for two of the five true positives. An important observation from the evaluation was that the model detected a very high proportion of sepsis patients in departments in which sepsis is not common. This was probably because septic patients differ more from the usual clientele than in, for example, emergency departments.
\par

\subsection{Limitations}
\subsubsection{Black box}
An important improvement in relation to clinical acceptance would be to implement supporting explanation methods in the predictions, such as layer-wise relevance propagation, deep Taylor decomposition, pattern attribution, or other DL explanation approaches \cite{Grgoire_Montavon_et_al_2017,Alexander_Binder_et_al_2016}. It is easy to imagine that a model that is more interpretable and supported by explanations would be more easily accepted in the clinic. Shickel et al. reached the same conclusion in a recent article reviewing the latest trends in the use of DL on EHR data \cite{Benjamin_Shickel_et_al_2017}. They completed their review with a warning against downplaying the importance of interpretability in favor of improvements in model performance.
\par

\subsubsection{Bias and confounding}
As the presented DL models (MLP and CNN-LSTM) operate in a high dimensional feature space not limited by domain specialists, it is important to consider the associated bias issues. In June 2018, Benjamin Recht and colleagues from UC Berkeley argued that many DL models may be less generalizable than we have assumed \cite{Benjamin_Recht_et_al_2018}. This claim was supported a month later by Zech et al., who showed that their DL models were significantly influenced by organizational and process-oriented elements \cite{John_R._Zech_et_al_2018}. Agniel et al. highlighted similar problems in a study on EHR data \cite{Denis_Agniel_et_al_2018}. The authors found that data regarding the time when the blood samples were ordered were more important than the blood test results for predicting three-year survival \cite{Jean-Louis_Vincent2016}. The important message in relation to sepsis detection based on EHR data is twofold: 1) If doctors or nurses have not measured certain vital signs or ordered certain blood samples, it will not be possible for models such as GB-Vital to predict sepsis. In this case, the CNN-LSTM model could still be used to estimate whether the patient is developing sepsis. 2) On the other hand, the CNN-LSTM model will most likely contain an unfortunate bias, which may be important if process-oriented elements change, such as new IT-systems or workflows. 
\par

\subsubsection{Reproducibility }
Another limitation of this study is that we do not test our models on the MIMIC-III database, unlike several of the studies we compare ourselves to. Due to the large differences in available data sources, it did not made sense to test our model on MIMIC-III data \cite{Alistair_E.W._Johnson_et_al_2016}. However, we suggest that our proposed methods should be tested on MIMIC-III data in the future.
\par

\subsubsection{Case-control matching}
In this study, we exclusively sampled our negative cases from simple naive rules, such as age and contact length. This means that our data sets potentially contain many patients that our algorithm could easily categorize as negatives. An improved sampling technique would be to match sepsis-positive contacts with "similar" sepsis-negative contacts in a case control matching approach, as suggested in \cite{Joseph_Futoma_et_al_2017AI}. In that study, the authors implemented a propensity scoring mindset that seemed to be inspired by causal inference estimation theory.
\par

\subsubsection{Dataset construction and oversampling}
In section \ref{inclusion_criteria_and_dataset}, we described how we oversampled the positive samples by a factor of 10 and then sampled negatives until we reached a ratio of 1:5. We explored many different combinations in relation to sampling techniques and balancing. Undersampling of the negative class worsened the test performance dramatically, indicating poor sampling of the variation space. Class ratios greater than 1:5 combined with weight-adjusted loss functions also reduced test performance, as did oversampling factors greater than ten. It could make sense to try more sophisticated sampling or data augmentation techniques to achieve better training performance.
\par

\section{Conclusion}
In this multi-center retrospective study, we present a novel deep learning system for early detection of sepsis in the heterogeneous data set present outside ICUs. The system learns representations of the key factors and interactions from the raw event sequence data itself, without relying on a labor-intensive feature extraction process. Our study indicates that sequential deep learning models can be used to detect sepsis at a very early stage, and we find that our model outperforms strong baseline models, such as GB-Vital, which rely on specific data elements and therefore suffer from many missing values in our data set. We also propose a new retrospective evaluation technique for assessing the clinical utility of the model that accounts for both intravenous antibiotics and blood culture requisitions. The evaluation showed that a large proportion of sepsis patients had not initiated intravenous antibiotics or blood culture at the time of early detection, and thus the model could facilitate such interventions at an earlier point in time.

Two interesting directions for future work would be to add supporting explanation methods into the predictions to improve clinical acceptance and to test our models on the MIMIC-III database.
\par
\section{Conflict of interest statement}
The authors Simon Meyer Lauritsen, Mads Ellersgaard Kal{\o}r, Emil Lund Kongsgaard and Bo Thiesson are employed at Enversion A/S.  

\section{Acknowledgements}
We acknowledge the steering committee for CROSS-TRACKS for access to data. For data acquisition, modeling, and validation, we thank the following: Julian Guldborg Birkemose,
Christian Bang, Per Dahl Rasmussen, Mathias Abitz Boysen, Anne Olsvig Boilesen, Lars Mellergaard,
and Jacob H{\o}y Berthelsen. For help with deep learning modeling, we thank Mads Kristensen. We
also thank the rest of the Enversion team for their support.\\
\noindent Funding: This work was supported by the Innovation Fond Denmark [case number 8053-00076B];
\section*{References}
\bibliography{sepsis}

\begin{thebibliography}{10}
\expandafter\ifx\csname url\endcsname\relax
  \def\url#1{\texttt{#1}}\fi
\expandafter\ifx\csname urlprefix\endcsname\relax\def\urlprefix{URL }\fi
\expandafter\ifx\csname href\endcsname\relax
  \def\href#1#2{#2} \def\path#1{#1}\fi

\bibitem{peterA2018}
A.~Perner, A.~T. Lassen, J.~Schierbeck, M.~Storgaard, N.~Reiter, T.~Benfield,
  Sygdomsbyrde og definitioner af sepsis hos voksne 180~(11).

\bibitem{SecretariaW2017}
W.~Secretariat,
  \href{http://apps.who.int/gb/ebwha/pdf_files/WHA70/A70_R7-en.pdf?ua=1}{Improving
  the prevention, diagnosis and clinical management of sepsis} (2017).
\newline\urlprefix\url{http://apps.who.int/gb/ebwha/pdf_files/WHA70/A70_R7-en.pdf?ua=1}

\bibitem{Murphy_SL_et_al_2013}
S.~Murphy, J.~Xu, K.~Kochanek, Deaths: final data for 2010. 61 (2013) 1--117.

\bibitem{Derek_C._Angus_et_al_2001}
D.~C. Angus, W.~T. Linde-Zwirble, J.~Lidicker, G.~Clermont, J.~Carcillo, M.~R.
  Pinsky,
  \href{http://dx.doi.org/10.1097/00003246-200107000-00002}{Epidemiology of
  severe sepsis in the united states: analysis of incidence, outcome, and
  associated costs of care} 29~(7) (2001) 1303--1310.
\newblock \href {http://dx.doi.org/10.1097/00003246-200107000-00002}
  {\path{doi:10.1097/00003246-200107000-00002}}.
\newline\urlprefix\url{http://dx.doi.org/10.1097/00003246-200107000-00002}

\bibitem{Jean-Louis_Vincent2016}
J.-L. Vincent, \href{http://dx.doi.org/10.1371/journal.pmed.1002022}{The
  clinical challenge of sepsis identification and monitoring} 13~(5) (2016)
  e1002022.
\newblock \href {http://dx.doi.org/10.1371/journal.pmed.1002022}
  {\path{doi:10.1371/journal.pmed.1002022}}.
\newline\urlprefix\url{http://dx.doi.org/10.1371/journal.pmed.1002022}

\bibitem{Jones_AE_et_al_2010}
A.~Jones, N.~Shapiro, S.~Trzeciak, R.~Arnold, H.~Claremont, J.~Kline,
  \href{https://dx.doi.org/10.1001/jama.2010.158}{Lactate clearance vs central
  venous oxygen saturation as goals of early sepsis therapy: a randomized
  clinical trial.} 303 (2010) 739--46.
\newblock \href {http://dx.doi.org/10.1001/jama.2010.158}
  {\path{doi:10.1001/jama.2010.158}}.
\newline\urlprefix\url{https://dx.doi.org/10.1001/jama.2010.158}

\bibitem{Joseph_Futoma_et_al_2017AI}
J.~Futoma, S.~Hariharan, M.~Sendak, N.~Brajer, M.~Clement, A.~Bedoya,
  C.~O'Brien, K.~Heller, \href{http://arxiv.org/abs/1708.05894v1}{An improved
  multi-output gaussian process {RNN} with real-time validation for early
  sepsis detection}, arXiv preprint arXiv:1708.05894.
\newline\urlprefix\url{http://arxiv.org/abs/1708.05894v1}

\bibitem{Steven_Horng_et_al_2017}
S.~Horng, D.~A. Sontag, Y.~Halpern, Y.~Jernite, N.~I. Shapiro, L.~A. Nathanson,
  \href{http://dx.doi.org/10.1371/journal.pone.0174708}{Creating an automated
  trigger for sepsis clinical decision support at emergency department triage
  using machine learning} 12~(4) (2017) e0174708.
\newblock \href {http://dx.doi.org/10.1371/journal.pone.0174708}
  {\path{doi:10.1371/journal.pone.0174708}}.
\newline\urlprefix\url{http://dx.doi.org/10.1371/journal.pone.0174708}

\bibitem{Joseph_Futoma_et_al_2017}
J.~Futoma, S.~Hariharan, K.~Heller,
  \href{http://arxiv.org/abs/1706.04152v1}{Learning to detect sepsis with a
  multitask gaussian process {RNN} classifier}, arXiv preprint
  arXiv:1706.04152.
\newline\urlprefix\url{http://arxiv.org/abs/1706.04152v1}

\bibitem{Qingqing_Mao_et_al_2018}
Q.~Mao, M.~Jay, J.~L. Hoffman, J.~Calvert, C.~Barton, D.~Shimabukuro, L.~Shieh,
  U.~Chettipally, G.~Fletcher, Y.~Kerem, Y.~Zhou, R.~Das,
  \href{http://dx.doi.org/10.1136/bmjopen-2017-017833}{Multicentre validation
  of a sepsis prediction algorithm using only vital sign data in the emergency
  department, general ward and {ICU}} 8~(1) (2018) e017833.
\newblock \href {http://dx.doi.org/10.1136/bmjopen-2017-017833}
  {\path{doi:10.1136/bmjopen-2017-017833}}.
\newline\urlprefix\url{http://dx.doi.org/10.1136/bmjopen-2017-017833}

\bibitem{Shamim_Nemati_et_al_2018}
S.~Nemati, A.~Holder, F.~Razmi, M.~D. Stanley, G.~D. Clifford, T.~G. Buchman,
  \href{http://dx.doi.org/10.1097/ccm.0000000000002936}{An interpretable
  machine learning model for accurate prediction of sepsis in the {ICU}} 46~(4)
  (2018) 547--553.
\newblock \href {http://dx.doi.org/10.1097/ccm.0000000000002936}
  {\path{doi:10.1097/ccm.0000000000002936}}.
\newline\urlprefix\url{http://dx.doi.org/10.1097/ccm.0000000000002936}

\bibitem{David_W_Shimabukuro_et_al_2017}
D.~W. Shimabukuro, C.~W. Barton, M.~D. Feldman, S.~J. Mataraso, R.~Das,
  \href{http://dx.doi.org/10.1136/bmjresp-2017-000234}{Effect of a machine
  learning-based severe sepsis prediction algorithm on patient survival and
  hospital length of stay: a randomised clinical trial} 4~(1) (2017) e000234.
\newblock \href {http://dx.doi.org/10.1136/bmjresp-2017-000234}
  {\path{doi:10.1136/bmjresp-2017-000234}}.
\newline\urlprefix\url{http://dx.doi.org/10.1136/bmjresp-2017-000234}

\bibitem{Calvert_JS_et_al_2016}
J.~Calvert, D.~Price, U.~Chettipally, C.~Barton, M.~Feldman, J.~Hoffman,
  M.~Jay, R.~Das,
  \href{https://dx.doi.org/10.1016/j.compbiomed.2016.05.003}{{A} computational
  approach to early sepsis detection.} 74 (2016) 69--73.
\newblock \href {http://dx.doi.org/10.1016/j.compbiomed.2016.05.003}
  {\path{doi:10.1016/j.compbiomed.2016.05.003}}.
\newline\urlprefix\url{https://dx.doi.org/10.1016/j.compbiomed.2016.05.003}

\bibitem{Md._Mohaimenul_Islam_et_al_2019}
M.~M. Islam, T.~Nasrin, B.~A. Walther, C.-C. Wu, H.-C. Yang, Y.-C. Li,
  \href{http://dx.doi.org/10.1016/j.cmpb.2018.12.027}{Prediction of sepsis
  patients using machine learning approach: {A} meta-analysis} 170 (2019) 1--9.
\newblock \href {http://dx.doi.org/10.1016/j.cmpb.2018.12.027}
  {\path{doi:10.1016/j.cmpb.2018.12.027}}.
\newline\urlprefix\url{http://dx.doi.org/10.1016/j.cmpb.2018.12.027}

\bibitem{Steve_Halligan_et_al_2015}
S.~Halligan, D.~G. Altman, S.~Mallett,
  \href{http://dx.doi.org/10.1007/s00330-014-3487-0}{Disadvantages of using the
  area under the receiver operating characteristic curve to assess imaging
  tests: {A} discussion and proposal for an alternative approach} 25~(4) (2015)
  932--939.
\newblock \href {http://dx.doi.org/10.1007/s00330-014-3487-0}
  {\path{doi:10.1007/s00330-014-3487-0}}.
\newline\urlprefix\url{http://dx.doi.org/10.1007/s00330-014-3487-0}

\bibitem{Kevin_McGeechan_et_al_2014}
K.~McGeechan, P.~Macaskill, L.~Irwig, P.~M. Bossuyt,
  \href{http://dx.doi.org/10.1186/1471-2288-14-86}{An assessment of the
  relationship between clinical utility and predictive ability measures and the
  impact of mean risk in the population} 14~(1).
\newblock \href {http://dx.doi.org/10.1186/1471-2288-14-86}
  {\path{doi:10.1186/1471-2288-14-86}}.
\newline\urlprefix\url{http://dx.doi.org/10.1186/1471-2288-14-86}

\bibitem{Talluri_R_Shete_S2016}
T.~R, S.~S, \href{https://dx.doi.org/10.1186/s12911-016-0336-x}{Using the
  weighted area under the net benefit curve for decision curve analysis.} 16
  (2016) 94.
\newblock \href {http://dx.doi.org/10.1186/s12911-016-0336-x}
  {\path{doi:10.1186/s12911-016-0336-x}}.
\newline\urlprefix\url{https://dx.doi.org/10.1186/s12911-016-0336-x}

\bibitem{Yann_LeCun_et_al_2015}
Y.~LeCun, Y.~Bengio, G.~Hinton,
  \href{http://dx.doi.org/10.1038/nature14539}{Deep learning} 521~(7553) (2015)
  436--444.
\newblock \href {http://dx.doi.org/10.1038/nature14539}
  {\path{doi:10.1038/nature14539}}.
\newline\urlprefix\url{http://dx.doi.org/10.1038/nature14539}

\bibitem{Morten_Schmidt_et_al_2015}
M.~Schmidt, S.~A.~J. Schmidt, J.~L. Sandegaard, V.~Ehrenstein, L.~Pedersen,
  H.~T. S{\o}rensen, \href{http://dx.doi.org/10.2147/clep.s91125}{The danish
  national patient registry: a review of content, data quality, and research
  potential} (2015) 449\href {http://dx.doi.org/10.2147/clep.s91125}
  {\path{doi:10.2147/clep.s91125}}.
\newline\urlprefix\url{http://dx.doi.org/10.2147/clep.s91125}

\bibitem{Carsten_Bcker_Pedersen2011}
C.~B. Pedersen, \href{http://dx.doi.org/10.1177/1403494810387965}{The danish
  civil registration system} 39~(7) (2011) 22--25.
\newblock \href {http://dx.doi.org/10.1177/1403494810387965}
  {\path{doi:10.1177/1403494810387965}}.
\newline\urlprefix\url{http://dx.doi.org/10.1177/1403494810387965}

\bibitem{Mitchell_M._Levy2003}
\href{http://dx.doi.org/10.1007/s00134-003-1662-x}{2001
  {SCCM/ESICM/ACCP/ATS/SIS} international sepsis definitions conference} 29~(4)
  (2003) 530--538.
\newblock \href {http://dx.doi.org/10.1007/s00134-003-1662-x}
  {\path{doi:10.1007/s00134-003-1662-x}}.
\newline\urlprefix\url{http://dx.doi.org/10.1007/s00134-003-1662-x}

\bibitem{Diederik_P._Kingma_Jimmy_Ba2014}
D.~P. Kingma, J.~Ba, Adam: A method for stochastic optimization, arXiv preprint
  arXiv:1412.6980.

\bibitem{Donahue_J_et_al_2017}
J.~Donahue, L.~Hendricks, M.~Rohrbach, S.~Venugopalan, S.~Guadarrama,
  K.~Saenko, T.~Darrell,
  \href{https://dx.doi.org/10.1109/TPAMI.2016.2599174}{Long-term recurrent
  convolutional networks for visual recognition and description.} 39 (2017)
  677--691.
\newblock \href {http://dx.doi.org/10.1109/TPAMI.2016.2599174}
  {\path{doi:10.1109/TPAMI.2016.2599174}}.
\newline\urlprefix\url{https://dx.doi.org/10.1109/TPAMI.2016.2599174}

\bibitem{Tara_N._Sainath_et_al_2015}
T.~N. Sainath, O.~Vinyals, A.~Senior, H.~Sak,
  \href{http://dx.doi.org/10.1109/icassp.2015.7178838}{Convolutional, long
  short-term memory, fully connected deep neural networks}, in: 2015 IEEE
  International Conference on Acoustics, Speech and Signal Processing (ICASSP),
  IEEE, 2015.
\newblock \href {http://dx.doi.org/10.1109/icassp.2015.7178838}
  {\path{doi:10.1109/icassp.2015.7178838}}.
\newline\urlprefix\url{http://dx.doi.org/10.1109/icassp.2015.7178838}

\bibitem{Alexis_Conneau_et_al_2017}
A.~Conneau, H.~Schwenk, L.~Barrault, Y.~Lecun,
  \href{http://dx.doi.org/10.18653/v1/e17-1104}{Very deep convolutional
  networks for text classification}, in: Proceedings of the 15th Conference of
  the European Chapter of the Association for Computational Linguistics: Volume
  1, Long Papers, Association for Computational Linguistics, 2017.
\newblock \href {http://dx.doi.org/10.18653/v1/e17-1104}
  {\path{doi:10.18653/v1/e17-1104}}.
\newline\urlprefix\url{http://dx.doi.org/10.18653/v1/e17-1104}

\bibitem{Alex_Graves2013}
A.~Graves, \href{http://arxiv.org/abs/1308.0850v5}{Generating sequences with
  recurrent neural networks}, arXiv preprint arXiv:1308.0850.
\newline\urlprefix\url{http://arxiv.org/abs/1308.0850v5}

\bibitem{Kevin_ten_Haaf_et_al_2017}
K.~ten Haaf, J.~Jeon, M.~C. Tammemägi, S.~S. Han, C.~Y. Kong, S.~K. Plevritis,
  E.~J. Feuer, H.~J. de~Koning, E.~W. Steyerberg, R.~Meza,
  \href{http://dx.doi.org/10.1371/journal.pmed.1002277}{Risk prediction models
  for selection of lung cancer screening candidates: {A} retrospective
  validation study} 14~(4) (2017) e1002277.
\newblock \href {http://dx.doi.org/10.1371/journal.pmed.1002277}
  {\path{doi:10.1371/journal.pmed.1002277}}.
\newline\urlprefix\url{http://dx.doi.org/10.1371/journal.pmed.1002277}

\bibitem{Andrew_J._Vickers_Angel_M._Cronin2010}
A.~J. Vickers, A.~M. Cronin,
  \href{http://dx.doi.org/10.1016/j.urology.2010.06.019}{Everything you always
  wanted to know about evaluating prediction models (but were too afraid to
  ask)} 76~(6) (2010) 1298--1301.
\newblock \href {http://dx.doi.org/10.1016/j.urology.2010.06.019}
  {\path{doi:10.1016/j.urology.2010.06.019}}.
\newline\urlprefix\url{http://dx.doi.org/10.1016/j.urology.2010.06.019}

\bibitem{E._W._Steyerberg_Y._Vergouwe2014}
E.~W. Steyerberg, Y.~Vergouwe,
  \href{http://dx.doi.org/10.1093/eurheartj/ehu207}{Towards better clinical
  prediction models: seven steps for development and an {ABCD} for validation}
  35~(29) (2014) 1925--1931.
\newblock \href {http://dx.doi.org/10.1093/eurheartj/ehu207}
  {\path{doi:10.1093/eurheartj/ehu207}}.
\newline\urlprefix\url{http://dx.doi.org/10.1093/eurheartj/ehu207}

\bibitem{Andrew_J._Vickers_Elena_B._Elkin2006}
A.~J. Vickers, E.~B. Elkin,
  \href{http://dx.doi.org/10.1177/0272989x06295361}{Decision curve analysis:
  {a} novel method for evaluating prediction models} 26~(6) (2006) 565--574.
\newblock \href {http://dx.doi.org/10.1177/0272989x06295361}
  {\path{doi:10.1177/0272989x06295361}}.
\newline\urlprefix\url{http://dx.doi.org/10.1177/0272989x06295361}

\bibitem{Valentin_Rousson_Thomas_Zumbrunn2011}
V.~Rousson, T.~Zumbrunn,
  \href{http://dx.doi.org/10.1186/1472-6947-11-45}{Decision curve analysis
  revisited: overall net benefit, relationships to {ROC} curve analysis, and
  application to case-control studies} 11~(1).
\newblock \href {http://dx.doi.org/10.1186/1472-6947-11-45}
  {\path{doi:10.1186/1472-6947-11-45}}.
\newline\urlprefix\url{http://dx.doi.org/10.1186/1472-6947-11-45}

\bibitem{Hilden_J_et_al_1978}
H.~J, H.~JD, B.~B, The measurement of performance in probabilistic diagnosis.
  {II.} trustworthiness of the exact values of the diagnostic probabilities. 17
  (1978) 227--37.

\bibitem{Steyerberg_EW_et_al_2010}
S.~EW, V.~AJ, C.~NR, G.~T, G.~M, O.~N, P.~MJ, K.~MW,
  \href{https://dx.doi.org/10.1097/EDE.0b013e3181c30fb2}{Assessing the
  performance of prediction models: a framework for traditional and novel
  measures.} 21 (2010) 128--38.
\newblock \href {http://dx.doi.org/10.1097/EDE.0b013e3181c30fb2}
  {\path{doi:10.1097/EDE.0b013e3181c30fb2}}.
\newline\urlprefix\url{https://dx.doi.org/10.1097/EDE.0b013e3181c30fb2}

\bibitem{Grgoire_Montavon_et_al_2017}
G.~Montavon, S.~Lapuschkin, A.~Binder, W.~Samek, K.-R. Müller,
  \href{http://dx.doi.org/10.1016/j.patcog.2016.11.008}{Explaining nonlinear
  classification decisions with deep taylor decomposition} 65 (2017) 211--222.
\newblock \href {http://dx.doi.org/10.1016/j.patcog.2016.11.008}
  {\path{doi:10.1016/j.patcog.2016.11.008}}.
\newline\urlprefix\url{http://dx.doi.org/10.1016/j.patcog.2016.11.008}

\bibitem{Alexander_Binder_et_al_2016}
A.~Binder, S.~Bach, G.~Montavon, K.-R. Müller, W.~Samek,
  \href{http://dx.doi.org/10.1007/978-981-10-0557-2_87}{Layer-wise relevance
  propagation for deep neural network architectures} (2016).
\newblock \href {http://dx.doi.org/10.1007/978-981-10-0557-2_87}
  {\path{doi:10.1007/978-981-10-0557-2_87}}.
\newline\urlprefix\url{http://dx.doi.org/10.1007/978-981-10-0557-2_87}

\bibitem{Benjamin_Shickel_et_al_2017}
B.~Shickel, P.~Tighe, A.~Bihorac, P.~Rashidi,
  \href{http://arxiv.org/abs/1706.03446v2}{Deep {EHR:} {A} survey of recent
  advances in deep learning techniques for electronic health record {(EHR)}
  analysis}, arXiv preprint arXiv:1706.03446\href
  {http://dx.doi.org/10.1109/JBHI.2017.2767063}
  {\path{doi:10.1109/JBHI.2017.2767063}}.
\newline\urlprefix\url{http://arxiv.org/abs/1706.03446v2}

\bibitem{Benjamin_Recht_et_al_2018}
B.~Recht, R.~Roelofs, L.~Schmidt, V.~Shankar, Do cifar-10 classifiers
  generalize to cifar-10?, arXiv preprint arXiv:1806.00451.

\bibitem{John_R._Zech_et_al_2018}
J.~R. Zech, M.~A. Badgeley, M.~Liu, A.~B. Costa, J.~J. Titano, E.~K. Oermann,
  \href{http://arxiv.org/abs/1807.00431v2}{Confounding variables can degrade
  generalization performance of radiological deep learning models}, arXiv
  preprint arXiv:1807.00431.
\newline\urlprefix\url{http://arxiv.org/abs/1807.00431v2}

\bibitem{Denis_Agniel_et_al_2018}
D.~Agniel, I.~S. Kohane, G.~M. Weber,
  \href{http://dx.doi.org/10.1136/bmj.k1479}{Biases in electronic health record
  data due to processes within the healthcare system: retrospective
  observational study} (2018) k1479\href {http://dx.doi.org/10.1136/bmj.k1479}
  {\path{doi:10.1136/bmj.k1479}}.
\newline\urlprefix\url{http://dx.doi.org/10.1136/bmj.k1479}

\bibitem{Alistair_E.W._Johnson_et_al_2016}
A.~E. Johnson, T.~J. Pollard, L.~Shen, L.~wei H.~Lehman, M.~Feng, M.~Ghassemi,
  B.~Moody, P.~Szolovits, L.~A. Celi, R.~G. Mark,
  \href{http://dx.doi.org/10.1038/sdata.2016.35}{{MIMIC-III,} a freely
  accessible critical care database} 3 (2016) 160035.
\newblock \href {http://dx.doi.org/10.1038/sdata.2016.35}
  {\path{doi:10.1038/sdata.2016.35}}.
\newline\urlprefix\url{http://dx.doi.org/10.1038/sdata.2016.35}

\end{thebibliography}
\end{document}